\documentclass[11pt]{article}

\usepackage[preprint]{acl}

\usepackage{times}
\usepackage{latexsym}

\usepackage[T1]{fontenc}

\usepackage[utf8]{inputenc}

\usepackage{microtype}

\usepackage{inconsolata}

\usepackage{graphicx}

\usepackage{subcaption}
\usepackage{booktabs} 
\usepackage{hyperref}

\usepackage{amsmath}
\usepackage{amssymb}
\usepackage{mathtools}
\usepackage{amsthm}

\usepackage[capitalize,noabbrev]{cleveref}

\usepackage{kotex}

\usepackage[table,xcdraw]{xcolor}
\usepackage{multirow}

\usepackage{xcolor}
\usepackage{soul}

\soulregister\ref7
\soulregister\pageref7
\soulregister\cref7
\soulregister\Cref7
\soulregister\autoref7
\soulregister\cite7
\soulregister\citep7
\soulregister\citet7
\soulregister\kbs0

\usepackage[textsize=tiny]{todonotes}

\usepackage{seqsplit}
\newcommand{\code}[1]{\texttt{\seqsplit{#1}}}

\usepackage{xspace}
\newcommand{\bench}{Korean Benchmarks\xspace}
\newcommand{\suite}{Korean Benchmark Suite\xspace}
\newcommand{\kbs}{KBS\xspace}

\title{\kbs: A \suite for Foundation Model Evaluation}

\author{
    \textbf{Yeonkyoung So\textsuperscript{1}}
    \textbf{Jongmin Kim\textsuperscript{1}}
    \textbf{Sungmok Jung\textsuperscript{1}}
    \textbf{Gyuseong Lee\textsuperscript{1}} \\
    \textbf{Sangho Kim\textsuperscript{1}}
    \textbf{Jongyeon Park\textsuperscript{1}}
    \textbf{Joonhak Lee\textsuperscript{1}}
    \textbf{Seho Pyo\textsuperscript{1}} \\
    \textbf{Gyeongje Cho\textsuperscript{1}}
    \textbf{Seorin Kim\textsuperscript{1}}
    \textbf{JiSoo Kim\textsuperscript{1}}
    \textbf{Suyoung Park\textsuperscript{1}} \\
    \textbf{Hyunji M. Park\textsuperscript{1}}
    \textbf{Yelim Ahn\textsuperscript{1}}
    \textbf{Yeongho Seo\textsuperscript{2}}
    \textbf{Jaejin Lee\textsuperscript{1,2}} \\ \\
\textsuperscript{1}Graduate School of Data Science, Seoul National University \\
\textsuperscript{2}Dept. of Computer Science and Engineering, Seoul National University \\[0.6em]
\begin{tabular}{c}
\texttt{\{kathy1028,jmin.kim,tjdahrwjd,ksnannaya,ksh4931,iscopark67,hmjelee,} \\
\texttt{hori14,avq159,seorin1116,jisoo.kim,itforspark,mhj233,mileya,} \\
\texttt{graysea2,jaejin\}@snu.ac.kr}
\end{tabular}
}

\begin{document}
\maketitle
\begin{abstract}
We present a \suite (\kbs) comprising Ko-ARC, Ko-GSM8K, Ko-EQ-Bench, Ko-WinoGrande, Ko-LAMBADA, and Ko-IFEval, covering six capabilities across 9,396 items. The suite combines review, localization, direct Korean construction, and redesign for Korean, avoiding a translation-only approach to benchmark adaptation. We evaluate Korean-developed models, multilingual open models, and closed API models under protocols that account for model access, prompt format, and scoring differences. Results show that Korean benchmark scores depend on task construction, scoring rules, prompting, and model post-training. These findings suggest that Korean evaluation should report both final scores and the settings used to obtain them.
\end{abstract}

\section{Introduction}

Large language models (LLMs) have rapidly advanced in recent years and are now used across a wide range of domains requiring language understanding, generation, and reasoning~\citep{brown2020language,sindhu2024evolution,zhou2025comprehensive}. As these models become more capable and more widely deployed, it becomes increasingly important to evaluate them using objective and standardized criteria~\citep{chang2024survey,desai2024emerging}. Benchmarks play a central role in this process by providing shared tasks for comparing models and tracking progress across capabilities such as knowledge, reasoning, contextual understanding, and instruction following~\citep{clark2018think,cobbe2021gsm8k,hendrycksmeasuring,sakaguchi2021winogrande, liang2022holistic,paech2023eqbench,zhou2023instructionfollowingevaluationlargelanguage,wang2024mmlu}. However, benchmark scores are only useful when the benchmark measures the intended capability rather than artifacts of dataset construction or evaluation~\citep{durmus-etal-2022-spurious,wu2022generating,bean2026measuring}. 

This issue is particularly important for languages outside a small set of major evaluation languages. English benchmarks have become broad and diverse enough that researchers can evaluate models on many capabilities using widely recognized test sets~\citep{srivastava2023beyond,shen2024taskbench}. For Korean, the situation has improved, and several valuable benchmarks have been introduced~\citep{park2klue,son2024hae,son2025kmmlu}. Still, it remains difficult to find Korean benchmark suites that both cover a broad range of capabilities and support close comparison with widely used English benchmarks. 

For this reason, Korean model evaluation often depends on adapting established English benchmarks. The most straightforward way to do this is direct translation, but translation-only evaluation can distort the benchmark itself. Literal translation, mistranslation, stylistic inconsistency, and cultural mismatch may change task difficulty, alter the naturalness of the prompt, or even shift what the task is testing~\citep{chuacrosslingual,doddapaneni2025cross,xu2025cross}. In such cases, differences in benchmark scores become harder to interpret, since they may reflect properties of the translation as much as the model's actual capability in Korean. 

In this work, we present a \suite (\kbs) for evaluating foundation models across multiple capabilities, including science question answering, mathematical reasoning, dialogue sentiment analysis, contextual reasoning, literary context understanding, and instruction following. Our construction pipeline combines machine translation with author-led review, linguistic correction, cultural localization, and cross-validation. For tasks that cannot be transferred appropriately by translation alone, we redesign the benchmark to preserve the intended capability in Korean rather than treating translation as a simple preprocessing step. 

Using the resulting benchmark suite, we evaluate both Korean LLMs and global LLMs with multilingual capability and examine how model behavior differs across tasks. For multiple-choice subsets, we also examine whether standard accuracy-based evaluation may be affected by superficial properties of the answer choices, which can make benchmark results harder to interpret~\citep{cho2026choices}. More broadly, our study suggests that benchmarking in languages beyond English requires careful consideration of both how benchmarks are transferred across languages and how evaluation protocols shape the interpretation of benchmark scores. 
The evaluation code and benchmark resources are publicly available.\footnote{Code: \url{https://github.com/mcrl/korean_benchmarks}.\\ Benchmark resources: \url{https://huggingface.co/collections/thunder-research-group/snu-thunder-llm-korean-benchmark-suite}.}
\section{Related Work}
\subsection{Benchmarking Foundation Models}
Benchmarks have become a standard tool for evaluating foundation models. By providing shared tasks and standardized metrics, they enable researchers to compare models across capabilities such as factual knowledge, reasoning, language understanding, code generation, and instruction following. Widely used examples include MMLU~\citep{hendrycksmeasuring} for multi-domain knowledge, ARC~\citep{clark2018think} for science question answering, GSM8K~\citep{cobbe2021gsm8k} for mathematical reasoning, WinoGrande~\citep{sakaguchi2021winogrande} and HellaSwag~\citep{zellers2019hellaswag} for commonsense and contextual reasoning, HumanEval~\citep{chen2021evaluating} for code generation, and IFEval~\citep{zhou2023instructionfollowingevaluationlargelanguage} for instruction following. 

Beyond individual task-specific datasets, holistic evaluation frameworks have attempted to aggregate diverse tasks into broader evaluation suites. HELM~\citep{liang2022holistic} evaluates models across multiple scenarios and metrics, while BIG-Bench~\citep{srivastava2023beyond} collects a large set of tasks designed to probe a wide range of model behaviors. As model performance has improved, more challenging benchmarks have also been introduced to better differentiate frontier models. Examples include MMLU-Pro~\citep{wang2024mmlu}, GPQA~\citep{rein2024gpqa}, and expert-level academic benchmarks~\citep{phan2025humanity}. Domain-specific benchmarks, such as LegalBench~\citep{guha2023legalbench} and MedQA~\citep{jin2021disease}, further show that general-purpose benchmarks are often insufficient for evaluating specialized capabilities. 

\subsection{Evaluation Sensitivity}
Benchmark scores are not always straightforward indicators of model capability. Recent work has shown that evaluation outcomes can depend on factors that are not directly related to the target skill being tested. For example, small changes in prompt wording or formatting can lead to different measured performance, suggesting that benchmark results may depend on the particular evaluation template used~\citep{sclar2023quantifying,polo2024efficient}. In multiple-choice question answering, model predictions can also be sensitive to the order, wording, length, format, or scoring of answer choices~\citep{zheng2023large,pezeshkpour2024large,wang2024look,molfese2025right}. More fundamentally, models may sometimes rely on the answer choices themselves, rather than on the question, so high accuracy does not always imply that the model solved the task in the intended way~\citep{balepur-etal-2024-artifacts,balepur2025these,wang2025llms,cho2026choices}.

These findings suggest that benchmark design involves more than selecting a task and reporting accuracy. The dataset construction process, prompt format, scoring rule, and evaluation protocol can all affect how benchmark scores should be interpreted. We account for these concerns in our multiple-choice evaluation by reporting NPSQ-based accuracy, which helps distinguish question-driven evidence from answer-choice artifacts. 

\begin{figure*}[htbp]
  \begin{center}
    \centerline{\includegraphics[width=\textwidth]{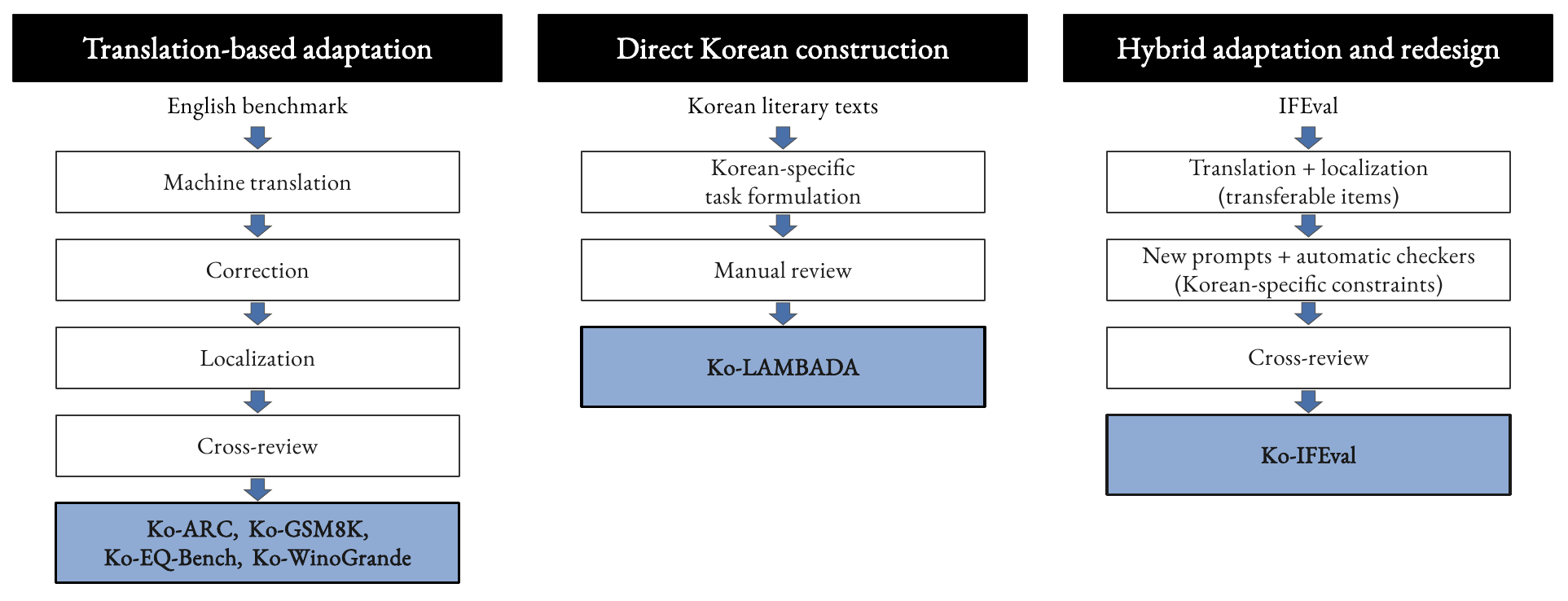}}
    \caption{
      Construction routes of the \suite. Translation-based tasks are adapted from English benchmarks through machine translation, correction, localization, and cross-review. Directly constructed tasks are built from Korean sources, while hybrid tasks combine translated items with Korean-specific prompts and automatic checkers. 
    }
    \label{fig:construction-routes}
  \end{center}
  \vskip -0.3in
\end{figure*}

\subsection{Korean LLM Evaluation}
Korean LLM evaluation has developed rapidly in recent years, but remains distributed across different goals, including general language understanding, cultural knowledge, professional knowledge, instruction following, and reasoning. Early resources such as KLUE, KorNLI/KorSTS, and KoBEST have provided important foundations for Korean NLU evaluation~\citep{ham2020kornli,park2klue,jang2022kobest}. KLUE has introduced a broad Korean language understanding benchmark, while KorNLI and KorSTS have adapted English NLI and STS resources into Korean through translation and human post-editing. KoBEST has further introduced Korean adaptations of several established tasks, motivated by the limitation that machine-translated benchmarks may not fully reflect Korean linguistic characteristics.

More recent benchmarks have expanded Korean evaluation toward native Korean sources and Korean-specific knowledge. HAE-RAE Bench evaluates Korean cultural and contextual knowledge~\citep{son2024hae}, while CLIcK draws from official Korean exams and textbooks to assess cultural and linguistic intelligence~\citep{kim2024click}. KMMLU provides large-scale expert-level multiple-choice questions collected from Korean exams~\citep{son2025kmmlu}, with later variants such as KMMLU-Redux and KMMLU-Pro focusing on question quality and professional knowledge~\citep{hong-etal-2025-kmmlu}. Other recent benchmarks target more specific dimensions, including national alignment and social values, Korean commonsense, and Korean mathematical reasoning~\citep{lee2024kornat,seo2024kocommongen,son2025multi,ko2025understand}. In addition, the Open Ko-LLM Leaderboard series has attempted to provide shared evaluation frameworks for Korean LLMs, but leaderboard-based efforts remain limited when the underlying datasets, evaluation details, or long-term maintenance are not fully available~\citep{park2024open,kim2025open}.

These efforts show that Korean LLM evaluation is no longer limited to a small number of datasets. They also show a clear shift from simply machine-translating English resources toward constructing tasks that better reflect Korean language use, culture, and real-world evaluation needs. 
Still, this expanding landscape raises a design question that is central to our work: how to determine when an English benchmark should be translated for comparability, localized for Korean naturalness, or redesigned because the original formulation no longer measures the same capability in Korean.

Our work addresses this question by constructing a multi-task Korean benchmark suite that combines reviewed translation, localization, and task redesign. We keep links to established English benchmarks where comparison is useful, but modify or reconstruct tasks when direct transfer would distort the intended evaluation. Together with the evaluation sensitivity considerations discussed above, this construction strategy supports more interpretable Korean benchmark results.
\section{Constructing \bench}
\label{3:construction}

\begin{figure*}[t]
  \begin{center}
    \centerline{\includegraphics[width=\textwidth]{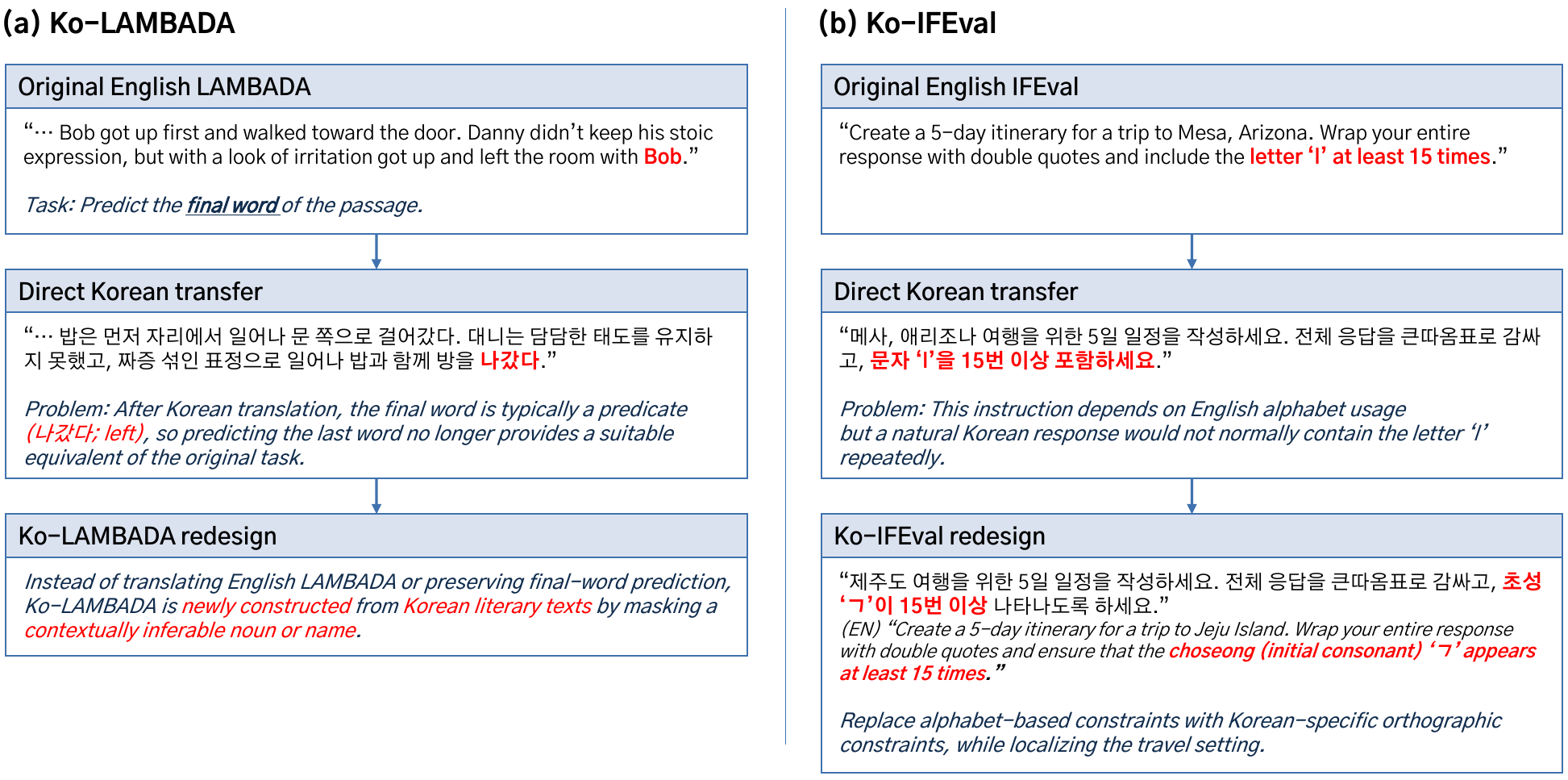}}
    \caption{
      Examples illustrating why task redesign is needed when adapting English benchmarks to Korean. Each panel contrasts direct transfer with the corresponding redesigned Korean task formulation.
    }
    \label{fig:redesign}
  \end{center}
  \vskip -0.3in
\end{figure*}

\subsection{Overview of the \suite}
\label{3_1:overview}

The \suite (\kbs) evaluates foundation models across six capabilities: science question answering, mathematical reasoning, dialogue sentiment analysis, commonsense and contextual reasoning, long-context literary understanding, and instruction following. It is built around established English benchmarks—ARC~\citep{clark2018think}, GSM8K~\citep{cobbe2021gsm8k}, EQ-Bench~\citep{paech2023eqbench}, WinoGrande~\citep{sakaguchi2021winogrande}, LAMBADA~\citep{paperno2016lambada}, and IFEval~\citep{zhou2023instructionfollowingevaluationlargelanguage}—while adapting each task to Korean according to its linguistic and cultural characteristics.

We construct the suite using three routes, as illustrated in Figure~\ref{fig:construction-routes}. First, Ko-ARC, Ko-GSM8K, Ko-EQ-Bench, and Ko-WinoGrande are adapted from their English counterparts through translation, correction, localization, and cross-review. Second, Ko-LAMBADA is directly constructed from Korean literary texts because the original English task formulation does not transfer cleanly to Korean. Third, Ko-IFEval follows a hybrid strategy: transferable instructions are translated and localized, while Korean-specific constraints and automatic checkers are newly developed. 

\begin{table*}[t]
\centering
\small
\begin{tabular}{llrll}
\toprule
\textbf{Benchmark} & \textbf{Capability} & \textbf{Items} & \textbf{Construction} & \textbf{English Source} \\
\midrule
Ko-ARC (easy)       & Science QA                         & 2,376 & Translated + Refined  & ARC-Easy \\
Ko-ARC (challenge)  & Science QA                         & 1,167 & Translated + Refined  & ARC-Challenge \\
Ko-GSM8K            & Mathematical reasoning              & 1,319 & Translated + Refined  & GSM8K \\
Ko-EQ-Bench         & Dialogue-based emotional inference  &   171 & Translated + Refined  & EQ-Bench \\
Ko-WinoGrande       & Commonsense/contextual reasoning    & 1,267 & Translated + Refined  & WinoGrande \\
Ko-LAMBADA          & Long-context literary understanding & 2,255 & Newly Constructed & LAMBADA \\
Ko-IFEval           & Instruction following               &   841 & Translated + Newly Constructed  & IFEval \\
\midrule
\textbf{Total}      &                                      & \textbf{9,396} & & \\
\bottomrule
\end{tabular}
\caption{
Composition of the \suite. Construction type indicates whether the benchmark is translated and refined from an English source, newly constructed from Korean sources, or combines translated items with newly constructed items for Korean-specific constraints. Ko-IFEval consists of 541 translated items and 300 newly constructed items.
}
\label{tab:stats}
\end{table*}


This task-level construction strategy allows us to retain comparability with established English benchmarks where appropriate, while redesigning tasks when direct transfer would change the capability being evaluated. The following subsections describe how English benchmarks are adapted to Korean, which tasks require redesign, and the resulting dataset statistics. 

\subsection{Adapting English Benchmarks to Korean}
\label{3_2:adapt}

For Ko-ARC, Ko-GSM8K, Ko-EQ-Bench, and Ko-WinoGrande, we adapt the corresponding English benchmarks through a four-stage pipeline: initial machine translation, author-led correction, localization, and independent cross-review. The goal is to preserve the original task structure and answer key where possible, while removing artifacts that would make the Korean version unnatural or misleading. Additional dataset-specific correction and localization rules are provided in Appendix~\ref{app:local}.

We first translate the English source items using the DeepL API\footnote{\url{https://www.deepl.com/products/api}}. We treat the machine-translated output only as a draft, since direct translation often produces literal phrasing, inconsistent terminology, and unnatural Korean expressions. Native Korean authors then revise each item, correcting both translation artifacts and issues in the original English items, such as typos, duplicate questions, incorrect answer labels, and incorrectly formatted solution annotations. For domain-specific tasks, authors also verify that scientific and mathematical terminology follows standard Korean usage and that the translated quantities, conditions, and explanations remain consistent with the answer. 

Localization is applied when a literal translation would make the item culturally unnatural or change the intended difficulty. We adjust names, units, currency, everyday objects, and culturally specific scenarios while preserving the causal or logical structure of the original item. For example, in math problems, we convert units and currencies to Korean conventions while keeping the arithmetic simple. We also replace culturally unfamiliar situations with Korean equivalents when the answer remains unchanged. For science questions, localization is applied selectively. We standardize Korean terminology and wording, but we do not replace foreign scientific examples simply to make them Korean when the original context is part of the scientific content. 

Finally, each item is independently cross-reviewed by an author who did not participate in the earlier correction and localization stage. This pass checks whether the Korean item is fluent, whether the answer key remains valid, and whether localization has introduced unintended ambiguity. If translation or localization makes the question ambiguous or changes the correct answer, we revise or remove the item. This final check reduces translation artifacts while keeping the Korean dataset aligned with the original benchmark.

\subsection{Task Redesign for Korean}
\label{3_3:redesign}

\begin{table*}[t]
\centering
\small
\begin{tabular}{p{2.5cm} p{5.5cm} p{2.3cm} p{4.8cm}}
\toprule
Category & Tasks & Models & Main Metrics \\
\midrule
Multiple-choice & Ko-ARC, Ko-WinoGrande, Ko-LAMBADA & Open only
& \texttt{acc} / \texttt{acc\_norm} / \texttt{acc\_npsq} \\
Generation & Ko-GSM8K, Ko-IFEval, Ko-EQ-Bench & Open + closed
& exact match / task-specific scores \\
Generative MCQA & Ko-ARC-gen, Ko-WinoGrande-gen, Ko-LAMBADA-gen & Open + closed
& exact match \\
\bottomrule
\end{tabular}
\caption{Summary of evaluation formats. Original multiple-choice evaluation is used for open-weight models with log-likelihood access, while closed API models are evaluated using generation-based variants.}
\label{tab:evaluation_tasks}
\end{table*}

The adaptation pipeline above is appropriate when an English task can be expressed in Korean without changing the capability being evaluated. For Ko-LAMBADA and Ko-IFEval, however, direct transfer is insufficient because the original task formulation depends partly on English-specific linguistic or formatting assumptions. We therefore redesign these tasks for Korean rather than treating translation as the default. Figure~\ref{fig:redesign} provides representative examples of these two redesign cases.

\paragraph{Ko-LAMBADA.}
Ko-LAMBADA is constructed directly from copyright-cleared Korean literary texts, primarily collected from Gongu Madang (공유마당)\footnote{\url{https://gongu.copyright.or.kr/gongu/main/main.do}}, instead of translating the original English LAMBADA.
We choose direct construction for two reasons. First, the original LAMBADA task asks models to predict the final word of an English passage. In many cases, this final word is a noun or name that must be inferred from the preceding context. In Korean, however, sentences typically end with predicates, so final-word prediction would often test verb endings rather than long-context understanding. Second, machine-translated literary passages often sound unnatural in Korean and tend to flatten literary expressions. We therefore construct new items from Korean literary sources rather than translating English literary passages.

To build candidate items, we automatically extract passage spans of an appropriate length from the collected Korean texts. We select spans that contain a target noun or name that can be masked, together with another in-context noun or name that can serve as a plausible distractor. The target is then masked in the passage, and candidate items are manually reviewed to ensure that the answer is inferable from context, that the distractor is plausible but not equally valid, and that the masked sentence remains grammatical and natural.

\paragraph{Ko-IFEval.}
Ko-IFEval follows a hybrid strategy. Some original IFEval instructions transfer naturally to Korean after translation and localization, such as constraints involving required keywords, forbidden words, or output format. Other constraints, on the other hand, depend on English-specific writing conventions, such as alphabet-based character constraints or word-count rules that do not map cleanly onto Korean morphology and spacing. For these cases, we add Korean-specific instruction types and adapt the automatic checkers accordingly. Examples include constraints based on Korean choseong (초성; leading consonant), Korean character-count rules, and formatting conventions that can be checked automatically for Korean outputs. As a result, Ko-IFEval keeps the original goal of evaluating instruction following with automatically checkable constraints, while extending the constraint set to Korean.

\subsection{Dataset Statistics}
\label{3_4:stat}

Table~\ref{tab:stats} summarizes the resulting benchmark suite. The \kbs contains 9,396 evaluation items across six capabilities. The largest subset is Ko-ARC, with 3,543 science question-answering items across easy and challenge subsets, followed by Ko-LAMBADA with 2,255 literary context-understanding items. Ko-GSM8K, Ko-WinoGrande, Ko-IFEval, and Ko-EQ-Bench contribute 1,319, 1,267, 841, and 171 items, respectively.
\section{Experiments}
\label{4:experiment}

\subsection{Evaluation Setup}
\label{4_1:setup}

\paragraph{Models.}
We evaluate both open-weight models and closed API models. Closed API models are evaluated through their official APIs: Claude Haiku 4.5 (\texttt{claude-haiku-4-5-20251001})~\citep{ANTHROPIC_2025} and Claude Opus 4.7 (\texttt{claude-opus-4-7})~\citep{ANTHROPIC_2026} through the Anthropic API\footnote{\url{https://platform.claude.com/docs/en/about-claude/models/overview}}, GPT-5-mini (\texttt{gpt-5-mini-2025-08-07}) and GPT-5.5 (\texttt{gpt-5.5-2026-04-23}) ~\citep{singh2025openai} through the OpenAI API\footnote{\url{https://developers.openai.com/api/docs}}. Open-weight models are evaluated with the HuggingFace Transformers~\citep{wolf-etal-2020-transformers} backend of Language Model Evaluation Harness (\texttt{lm-evaluation-harness})~\citep{biderman2024lessons}. All open-model runs use bfloat16 precision on a single NVIDIA B200 GPU.

We evaluate two groups of open-weight models: Korean-developed LLMs spanning five families (KT Mi:dm 2.0~\citep{KT_2025}, LG EXAONE 4.0~\citep{research2025exaone1}, Kakao Kanana-2~\citep{kanana2_blog}, Naver HyperCLOVAX~\citep{naver2025hyperclova}, and SKT A.X 4.0~\citep{SKT_2025}) and widely used multilingual LLMs (Qwen 3.5~\citep{qwen35_blog} and Gemma 4~\citep{gemma4_model_card}). This model set allows direct comparison between Korean-specialized and multilingual model families. The full list of evaluated open-weight models, parameter sizes, and HuggingFace model identifiers is provided in Appendix~\ref{app:models}.

\paragraph{Evaluation formats.}
We use log-likelihood-based multiple-choice evaluation for open-weight models on Ko-ARC, Ko-WinoGrande, and Ko-LAMBADA. For the evaluated closed API models, token-level log-likelihoods are unavailable. To compare the two model groups under a common response format, we additionally evaluate both groups on generative variants of these tasks.
Models are presented with explicitly labeled answer options and instructed to output a single label. Ko-ARC uses A-D labels, while Ko-WinoGrande and Ko-LAMBADA use A-B labels. The final answer is extracted from the generated response with a regular expression over the option labels.

\paragraph{Diagnostic translation variants.}
We additionally evaluate English and raw DeepL variants to examine performance differences associated with evaluation language and author-led correction and localization. These comparisons are restricted to variants with meaningful item-level correspondence. The DeepL variant uses the raw machine-translated draft before correction and is available for Ko-ARC, Ko-GSM8K, and Ko-EQ-Bench. The English variant uses the corresponding original English items. For Ko-WinoGrande, we report only the \kbs-English comparison because its Korean options were manually constructed and the sentence was revised accordingly, leaving no full raw-DeepL counterpart.

\paragraph{Prompt formatting.}
For open-weight models, we use plain prompting for all models and additionally evaluate chat-template prompting for instruction-tuned or chat-oriented models with a defined chat template. Closed API models are evaluated only with chat-style prompting, since their official APIs do not provide an equivalent plain prompting interface.

\paragraph{Generation settings.}
All generation-based evaluations use greedy decoding. This includes Ko-GSM8K, Ko-IFEval, Ko-EQ-Bench, the generative multiple-choice variants evaluated for both open-weight and closed API models.
The model always selects the highest-probability next token rather than sampling from the output distribution. We disable sampling with \texttt{do\_sample=False} and set the temperature to 0.0. Each reported result is based on a single run. 

\paragraph{Metrics.}
Table~\ref{tab:evaluation_tasks} summarizes the task formats and metrics. For multiple-choice tasks, we report accuracy under raw log-likelihood (\texttt{acc}), byte-length normalized log-likelihood (\texttt{acc\_norm}), and NPSQ-based accuracy (\texttt{acc\_npsq}). NPSQ (\textit{Normalized Probability Shift by the Question}) compares the model's preference among answer choices conditioned on the question against its preference computed without the question, 
and selects the choice with the largest normalized increase in log-probability when the question is added.
This separates question-driven evidence from superficial choice preference that arises from lexical, length, or surface-form biases over the answer choices alone~\citep{cho2026choices}. 

For generation tasks, we report task-specific automatic metrics, such as answer exact match for Ko-GSM8K, strict and loose instruction-following accuracy for Ko-IFEval, and the custom Ko-EQ-Bench score. The full evaluation protocol, including stop sequences, prompting templates, answer-extraction rules, and metric definitions, is provided in Appendix~\ref{app:evaluation_protocol}.

\subsection{Results}
\label{4_2:result}


\paragraph{Overall performance.}
Among open-weight models under default chat prompting, Gemma-4-31B-it leads Ko-ARC and Ko-WinoGrande under NPSQ, and achieves the highest Ko-GSM8K strict exact match (82.71\%) and Ko-IFEval prompt-level strict accuracy (85.02\%). On Ko-LAMBADA, EXAONE-4.0-32B achieves the highest NPSQ accuracy (83.68\%). The leading models change under plain prompting: Gemma-4-31B leads Ko-ARC and Ko-WinoGrande, Kanana-2 variants lead Ko-LAMBADA and Ko-GSM8K, and A.X 4.0 Light leads Ko-IFEval. Complete results are provided in Appendix~\ref{app:full-results}.


\begin{table}[htbp]
\centering
\resizebox{0.8\columnwidth}{!}{%
\begin{tabular}{@{}lrrr@{}}
\toprule
\multicolumn{1}{c}{\textbf{Model group}} &
\multicolumn{1}{c}{\textbf{English}} &
\multicolumn{1}{c}{\textbf{DeepL}} &
\multicolumn{1}{c}{\textbf{\kbs}} \\
\midrule
\textbf{Korean-developed} & 67.29 & 54.00 & 59.49 \\
\textbf{Qwen3.5}          & 51.22 & 44.14 & 46.49 \\
\textbf{Gemma 4}          & 72.56 & 60.14 & 67.05 \\
\bottomrule
\end{tabular}%
}
\caption{
ARC accuracy (\%) using NPSQ under chat prompting. All versions are restricted to the same 2,376 Easy and 1,167 Challenge items. Scores are averaged equally over the two subsets
and then over the models in each group. DeepL denotes the raw machine-translated version before author correction and localization.
}
\label{tab:translation_language_effects}
\end{table}

\paragraph{Translation and language effects.}
We use ARC for this comparison because it provides matched English, raw DeepL, and \kbs variants under the same log-likelihood MCQA setup. Table~\ref{tab:translation_language_effects} compares the original English items, the raw DeepL translations before review, and the finalized \kbs items using open-weight chat-template evaluations with \texttt{acc\_npsq}, restricted to items shared by all three versions. Models score higher on \kbs than on raw DeepL in all three model groups, with gains of 2.35-6.91 percentage points. These gains suggest that correcting translation errors and adapting questions to Korean contexts help models better understand the intended meaning of each question. Models generally score higher on the English original than on the Korean variants on ARC. However, the English-Korean gap is task-dependent: on WinoGrande, where only the \kbs-English comparison is meaningful, the gap is much smaller (Table~\ref{tab:language_comparison_appendix}).

\paragraph{Scoring differences in MCQA tasks.}
\begin{table}[htbp]
\centering
\resizebox{\columnwidth}{!}{%
\begin{tabular}{@{}lcrrr@{}}
\toprule
\multicolumn{1}{c}{\textbf{Task}} & \textbf{Prompt format} & \multicolumn{1}{c}{\textbf{acc}} & \multicolumn{1}{c}{\textbf{acc\_norm}} & \multicolumn{1}{c}{\textbf{acc\_npsq}} \\ \midrule\midrule
\multirow{2}{*}{Ko-WinoGrande} & plain & 61.69 & 61.06 & 59.33 \\
 & chat & 58.08 & 57.64 & 55.73 \\ \midrule
\multirow{2}{*}{Ko-LAMBADA} & plain & 92.49 & 92.13 & 87.39 \\
 & chat & 81.22 & 80.06 & 69.20 \\ \midrule
\multirow{2}{*}{Ko-ARC-Easy} & plain & 67.81 & 69.04 & 71.21 \\
 & chat & 64.33 & 62.04 & 64.15 \\ \midrule
\multirow{2}{*}{Ko-ARC-Challenge} & plain & 45.63 & 51.57 & 52.77 \\
 & chat & 44.53 & 50.03 & 49.52 \\ \bottomrule
\end{tabular}%
}
\caption{Mean multiple-choice accuracy of open-weight models under three scoring rules: raw log-likelihood (\texttt{acc}), byte-length normalized log-likelihood (\texttt{acc\_norm}), and NPSQ-based accuracy (\texttt{acc\_npsq}).}
\label{tab:mcqa-scoring}
\end{table}
Table~\ref{tab:mcqa-scoring} shows that the effect of the scoring rule is task-dependent. NPSQ-based accuracy is lower than raw accuracy on Ko-LAMBADA, especially under chat-template prompting, where the mean score drops from 81.22 to 69.20. Ko-WinoGrande shows a smaller decrease under NPSQ. In contrast, both Ko-ARC subsets often obtain higher scores under \texttt{acc\_npsq} than under raw accuracy, with the larger difference on Ko-ARC-Challenge. Byte-length normalization has its largest effect on Ko-ARC-Challenge. These results suggest that answer-choice priors affect Ko-LAMBADA most strongly, while the impact of scoring rules varies with the structure of each multiple-choice task. 

\begin{figure*}[h]
  \begin{center}
    \centerline{\includegraphics[width=0.95\textwidth]{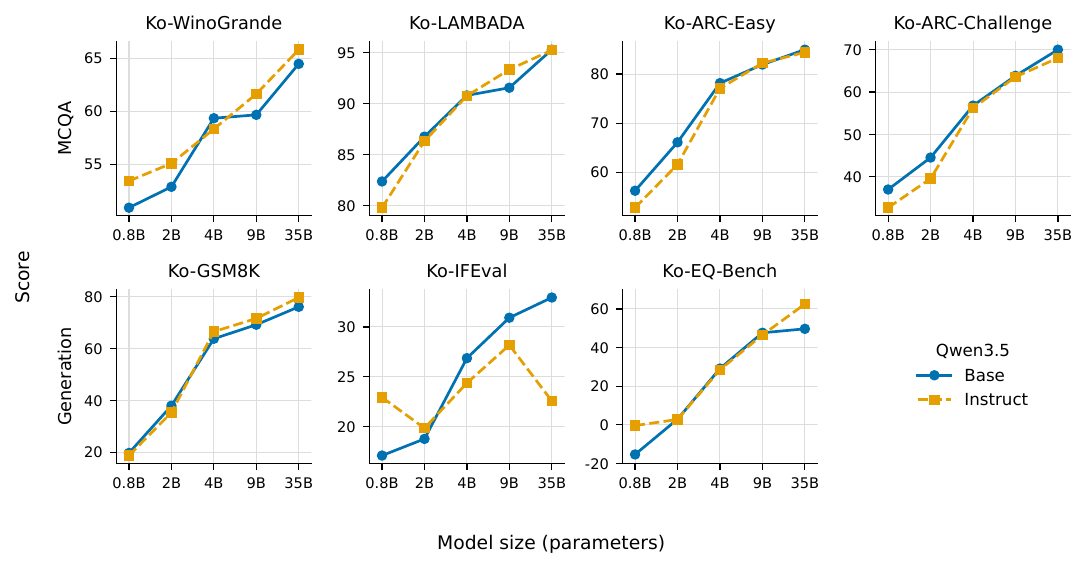}}
    \caption{
      Performance of Qwen3.5 Base (pretrained) and instruction-tuned variants at five paired model sizes under plain prompting. We report NPSQ accuracy for MCQA, strict exact match for Ko-GSM8K, prompt-level strict accuracy for Ko-IFEval, and the EQ-Bench score for Ko-EQ-Bench. Accuracies are expressed as percentages. 35B denotes 35B-A3B.
    }
    \label{fig:qwen_size_instruction}
  \end{center}
  \vskip -0.3in
\end{figure*}

\paragraph{Effects of prompt formatting.}

Prompt format has a large effect on open-weight models, but the direction of the effect differs by model family (Appendix~\ref{app:full-results} reports the complete results, including plain prompting averages over the chat-template subset). For several chat models, applying the chat template substantially improves generation performance over plain prompting. For example, Gemma 4 and EXAONE-4.0 model families show large gains under chat prompting across Ko-GSM8K, Ko-IFEval, and Ko-EQ-Bench. These cases highlight that evaluations without the intended chat template can substantially underestimate the capabilities of instruction-tuned models.


\begin{table}[t]
\centering
\small
\setlength{\tabcolsep}{3pt}
\begin{tabular}{@{}lrrrr@{}}
\toprule
& \multicolumn{2}{c}{Ko-GSM8K}
& \multicolumn{2}{c}{Ko-EQ-Bench} \\
\cmidrule(lr){2-3}\cmidrule(lr){4-5}
Qwen3.5 & Default & Off & Default & Off \\
\midrule
0.8B    & 23.05 & 23.96 & 16.77 & 16.65 \\
2B      & 43.06 & 44.50 & 31.16 & 30.98 \\
4B      &  0.15 & 69.60 &  0.00 & 53.11 \\
9B      &  0.00 & 77.33 &  0.00 & 54.55 \\
27B     &  0.00 & 80.67 &  0.00 & 72.47 \\
35B-A3B &  0.08 & 80.67 &  0.00 & 70.45 \\
\bottomrule
\end{tabular}
\caption{
Qwen3.5 results under default chat prompting and with thinking disabled (Off). Ko-GSM8K reports strict exact match (\%). Ko-EQ-Bench reports the EQ-Bench score. Both conditions use chat templates.
}
\label{tab:qwen_thinking}
\end{table}

The reverse pattern appears in the Qwen3.5 series. Models larger than 2B perform well on Ko-GSM8K under plain prompting but drop to almost zero under chat prompting, and flexible answer extraction does not recover the scores. Output inspection suggests that the chat template leads these larger models to produce long English reasoning traces, often without reaching a final Korean answer (Appendix~\ref{app:qwen-traces}). Prior work has also reported English-centric representations and language inconsistency in multilingual reasoning~\citep{schut2025multilingual, zhang2025think}. We additionally evaluate six instruction-tuned Qwen3.5 models with thinking disabled through \texttt{enable\_thinking=False}, while retaining chat templates. As Table~\ref{tab:qwen_thinking} shows, the four larger models recover to 69.60-80.67\% strict exact match on Ko-GSM8K and 53.11-72.47 on Ko-EQ-Bench. These results indicate that enabling thinking does not necessarily improve performance on these Korean tasks under the evaluated generation settings. The recovery with thinking disabled highlights the need to account for reasoning mode when interpreting benchmark scores.

A different pattern appears for some Korean-developed models. Mi:dm 2.0 Base and Kanana-2-30b-a3b-instruct-2601 obtain much higher flexible extract than strict match scores on Ko-GSM8K under chat prompting, indicating that many outputs contain the correct numerical answer but do not follow the expected final answer format. In these cases, the low strict score reflects answer format mismatch rather than complete reasoning failure. Taken together, these results show that prompt format is not a minor implementation detail. It interacts with the model's post-training and can change whether the model follows the intended task, switches to English reasoning on Korean inputs, or fails the required answer format.

\paragraph{Effects of instruction tuning and model size.}
Figure~\ref{fig:qwen_size_instruction} compares Qwen3.5 Base and instruction variants across model sizes under plain prompting. Larger models generally perform better, especially on Ko-ARC and Ko-GSM8K. However, instruction tuning is not a uniform gain in multilingual model families. In Qwen3.5, the instruction variant improves some scores, such as Ko-GSM8K at larger sizes, but does not consistently outperform the Base model across MCQA tasks or Ko-IFEval. A similar pattern appears in Gemma under plain prompting, where instruction-tuned models often underperform their Base counterparts. On the other hand, the Kanana series shows progressive improvements on Ko-GSM8K and Ko-IFEval under plain prompting, with the Instruct variant also achieving the highest Ko-EQ-Bench score among the three variants.
These results suggest that instruction tuning effects depend on what the post-training process is optimized for, including the target language, task format, and prompting style.

\paragraph{Closed API models.}

Table~\ref{tab:generative_mcqa_comparison} compares the best open-weight and closed API scores under a common generative MCQA format with default chat configurations. The best closed scores exceed the best open-weight scores by 1.06-2.62 percentage points on Ko-ARC and Ko-LAMBADA, while the gap is larger on Ko-WinoGrande (8.52 points). With thinking disabled, Qwen3.5-27B achieves 98.53\% on Ko-ARC-Easy, further narrowing the gap to 0.59 points.

The relative performance also varies across generation tasks. On Ko-GSM8K, GPT-5.5 and Claude Opus 4.7 achieve 96.29\% and 96.06\% strict exact match, respectively, compared with 82.71\% for Gemma-4-31B-it, the best open-weight model under default chat prompting. On Ko-IFEval, however, Gemma-4-31B-it achieves 85.02\% prompt-level strict accuracy, exceeding the best closed score of 81.09\% from GPT-5.5. Complete results are provided in Appendix~\ref{app:full-results}.

\begin{table}[t]
\centering
\small
\setlength{\tabcolsep}{4pt}
\begin{tabular}{@{}lrrr@{}}
\toprule
Task & Best open & Best closed & Gap \\
\midrule
Ko-ARC-Easy      & 98.06 & 99.12 & 1.06 \\
Ko-ARC-Challenge & 96.57 & 97.77 & 1.20 \\
Ko-LAMBADA       & 96.14 & 98.76 & 2.62 \\
Ko-WinoGrande    & 84.85 & 93.37 & 8.52 \\
\bottomrule
\end{tabular}
\caption{
Best generative MCQA accuracy (\%) among open-weight and closed API models under default chat configurations. Gemma-4-31B-it leads the open-weight group on all four tasks. GPT-5.5 leads the closed group on Ko-ARC-Challenge and Claude Opus 4.7 on the other tasks. Gap denotes the accuracy difference (closed minus open), in percentage points. Qwen non-thinking runs are reported separately in Appendix~\ref{app:full-results}.
}
\label{tab:generative_mcqa_comparison}
\end{table}
\section{Conclusion}
We present a \suite for evaluating foundation models across six capabilities. The suite is constructed through task-level routes that combine reviewed translation and localization, direct Korean construction, and Korean-specific redesign. To our knowledge, few Korean evaluation suites jointly address broad capability coverage, task-specific benchmark construction, differences in model access, and prompt-formatting conditions. For multiple-choice tasks, we use log-likelihood-based scoring and NPSQ-based accuracy for open-weight models, and evaluate closed API models via label generation. For generation tasks, both open-weight and closed-API models are evaluated based on model-generated outputs. We further compare plain prompting and chat-template prompting for open-weight models, where applicable, while closed-API models are evaluated under chat-style prompting. The results show that Korean benchmark scores are shaped by task construction, scoring rules, prompt format, and model post-training choices. These factors can change both the measured score and the interpretation of model behavior. We therefore suggest that Korean evaluations report not only final scores but also how each task is adapted, how each score is computed, and under which prompting and model-access settings the score is obtained.

\section*{Limitations}
Although the \suite covers six capabilities, it does not represent the full range of Korean language use, such as long-form generation, multi-turn dialogue, or safety-related behavior. Moreover, some tasks are adapted from English benchmarks, so residual translation or localization artifacts may remain despite author-led review. In addition, our evaluation relies on automatic metrics and single-run greedy decoding, which may not capture all aspects of model behavior. Future work should evaluate robustness across broader prompting and decoding conditions.

\kbs may be affected by contamination from publicly available source material. The English-Korean score gaps suggest sensitivity to evaluation language and adaptation, but do not rule out contamination.
\section*{Ethical Considerations} 
This work aims to improve the reliability and reproducibility of Korean foundation model evaluation by providing a benchmark suite with documented construction and evaluation protocols. We expect the suite to support more transparent comparison of Korean and multilingual models, and to make evaluation resources for Korean more accessible to the research community. We caution that benchmark scores should be interpreted only as reflecting performance under the specific tasks, prompts, scoring rules, and evaluation settings described in this paper.

We have released the evaluation code under the MIT license. Dataset subsets adapted from existing English benchmarks have been released under the licenses of their respective source datasets, including CC-BY for WinoGrande, CC-BY-SA for ARC, MIT for GSM8K and EQ-Bench, and Apache 2.0 for IFEval. Ko-LAMBADA, which is newly constructed from copyright-cleared Korean literary texts, has been released under CC-BY-NC-SA 4.0.
\section*{Acknowledgments}
This work was partially supported by the National Research Foundation of Korea (NRF) under Grant No. RS-2023-00222663 (Center for Optimizing Hyperscale AI Models and Platforms) and under Grant No. A400-20260031, and by the Institute for Information and Communications Technology Promotion (IITP) under Grant No. 2018-0-00581 (CUDA Programming Environment for FPGA Clusters) and No. RS-2025-02304554 (Efficient and Scalable Framework for AI Heterogeneous Cluster Systems), all funded by the Ministry of Science and ICT (MSIT) of Korea. It was also partially supported by the Korea Health Industry Development Institute (KHIDI) under Grant No. RS-2025-25454559 (Frailty Risk Assessment and Intervention Leveraging Multimodal Intelligence for Networked Deployment in Community Care), funded by the Ministry of Health and Welfare (MOHW) of Korea. Additional support was provided by the BK21 Plus Program for Innovative Data Science Talent Education (Department of Data Science, Seoul National University, No. 5199990914569) and the BK21 FOUR Program for Intelligent Computing (Department of Computer Science and Engineering, Seoul National University, No. 4199990214639), both funded by the Ministry of Education (MOE) of Korea. This work was also partially supported by the Advanced GPU Utilization Support Program, funded by the Ministry of Science and ICT (MSIT) of Korea and operated by the National IT Industry Promotion Agency (NIPA). Research facilities were provided by the Institute of Computer Technology (ICT) at Seoul National University.

\bibliography{custom}

\appendix




\section{Detailed Correction and Localization Rules}
\label{app:local}

This section describes the dataset-specific rules used during the author-led correction and localization stage introduced in Section~\ref{3_2:adapt}. Throughout these stages, we aim to preserve the original reasoning requirements and difficulty, while verifying that the answer remains valid after localization.

\subsection{Source-Side Corrections in English Benchmarks}
\label{app:source-error}

Some inconsistencies are present in the English source benchmarks rather than being introduced by translation. We correct these cases during the correction and localization stage so that the Ko-\textsc{X} dataset reflects the intended task design of the corresponding \textsc{X} dataset rather than source-side artifacts. Representative cases include:

\begin{itemize}
\item \textbf{ARC-Challenge near-duplicates.} The items \texttt{LEAP\_8\_10365} and \texttt{LEAP\_2000\_8\_2} differ only in the surface form "one centimeter" vs. "1 centimeter". We retain one item and remove the other. Other near-duplicate pairs are handled using the same criterion.
\item \textbf{ARC-Easy unintended numeral.} The question stem "Which of the 27 following is the best use of a robot?" in \texttt{MCAS\_2005\_8\_27} contains an unintended "27", which we remove.
\item \textbf{GSM8K calculator annotations.} GSM8K marks calculator-evaluable expressions with \texttt{<<...>>}. Redundant annotations such as \texttt{<<8=8>>} are removed, and annotations with incorrect arithmetic are corrected. The original annotation convention is otherwise preserved for compatibility with existing GSM8K evaluation code.
\end{itemize}

\subsection{Ko-GSM8K: Names, Units, and Currency}
\label{app:gsm8k-rules}

\paragraph{Names.}
English given names are replaced with Korean given names sampled from the statistics of high-frequency birth names recorded between 2008 and 2024 in the Korean Supreme Court e-family Register\footnote{\url{https://stfamily.scourt.go.kr/st/StFrrStatcsView.do?pgmId=090000000025}}. Name substitutions are kept consistent within each problem so that the same person is referred to by the same Korean name throughout the problem. 

\paragraph{Units.}
Imperial units are converted to metric units. When a numeric value is part of the arithmetic structure of the problem, we use an arithmetically simple metric counterpart rather than an exact conversion (e.g., 10 feet $\rightarrow$ 3 meters rather than 3.048 meters). After each conversion, we recompute the answer to verify that the localized item preserves the intended difficulty.

\paragraph{Currency.}
U.S. dollars are converted to Korean won using the fixed rate 1 USD = 1,000 KRW. This rate is chosen to preserve the arithmetic structure of the original problems while keeping the quantities natural in Korean. For example, if a problem asks for the cost of three items priced at \$15 each, the original calculation is simply \(15 \times 3 = 45\). Using a realistic exchange rate such as 1 USD = 1,512 KRW would turn this into \(15 \times 1512 \times 3\), adding unintended arithmetic complexity. With the fixed rate, the problem becomes \(15000 \times 3 = 45000\), preserving the intended multiplication difficulty. Although this abstracts away from exact exchange-rate variation, the rounded conversion improves arithmetic clarity.

\paragraph{Korean counting expressions.}
Korean counting expressions are selected according to the noun being counted. For example, "5 pencils" is translated as "연필 5자루", where "자루" is the counting word used with pencil-like objects. We apply the same principle to other noun classes, including people, books, vehicles, and animals, etc. 

\paragraph{Everyday items, brands, and places.}
Everyday items, foods, brands, and place names that are less natural in Korean contexts are replaced with familiar Korean counterparts, provided that the substitution does not change the mathematical structure of the original problem. 

\subsection{Ko-ARC: Cultural and Geographic Localization}
\label{app:arc-rules}

ARC items often contain U.S.-specific contextual details, such as states, national parks, coastlines, animal species, and industrial settings. When these details are not essential to the scientific concept being tested, we replace them with Korean or East Asian counterparts that are more natural in Korean educational contexts. Each substitution is checked against relevant Korean educational and public reference materials. Table~\ref{tab:arc-localize} summarizes representative substitution rules.

\begin{table*}[t]
\centering
\small
\begin{tabular}{p{3.2cm} p{4.6cm} p{6.8cm}}
\toprule
\textbf{Category} & \textbf{Original context in English ARC} & \textbf{Localized context in Ko-ARC} \\
\midrule
Locations &
U.S. states, national parks, coastlines, and rivers &
Korean or East Asian locations with analogous phenomena (e.g., U.S.\ Gulf coast~$\to$~West Sea / Yellow Sea; Maine solar-noon example~$\to$~Baengnyeongdo Island). Globally familiar references, such as the Himalayas, are retained when they are already standard in Korean science education. \\

Climate and seasonality &
Hurricanes; jet-stream examples set in the continental U.S. & Comparable weather and seasonal phenomena in the Northwest Pacific (e.g., \emph{hurricanes}~$\to$~\emph{typhoons}; jet-stream examples rewritten using Korean winter-weather patterns; daylight-duration comparisons are rewritten using Korean cities with appropriate latitude differences). \\

Species and organisms &
Black bear; warbler migration; Gulf sturgeon & Ecologically comparable species that are familiar in Korean educational contexts (e.g., Asiatic black bear in Jirisan Mountain; cranes migrating between Siberia or Manchuria and the Korean peninsula; salmon as a river-sea migratory species). \\

Industry and energy &
Kentucky coal-transport context; U.S.\ 2003 electricity generation mix &
Korean settings with comparable industrial or energy-related roles (e.g., Jecheon-si cement freight corridor; Korean 2022 electricity generation mix). \\

Units and measurements &
Imperial units; city-by-city daylight comparisons &
Metric units; numerical values adjusted when needed to preserve the scientific relation; city pairs reselected within Korea. \\

Everyday objects and cultural examples &
U.S.-specific household or cultural items used to frame the question &
Korean counterparts that make the question natural in Korean while preserving the scientific framing and the correct answer. \\
\bottomrule
\end{tabular}
\caption{Representative localization rules for Ko-ARC. Each substitution is checked to preserve both the underlying scientific mechanism and the correct answer.}
\label{tab:arc-localize}
\end{table*}

\subsection{Ko-EQ-Bench: Emotion Labels and Honorific Usage}
\label{app:eqbench-rules}

\paragraph{Emotion-label mapping.}
Some English emotion labels encode distinctions that do not map cleanly onto a single Korean label. To ensure consistent annotation across items, we use a documented mapping table for emotion labels. When a direct label mapping does not preserve the intended meaning in context, we apply an item-specific revision. For example, "Victimized" is mapped to "피해의식", which captures a perceived sense of "being wronged" rather than "the literal state of having been harmed". Idiomatic expressions are also translated by preserving their contextual meaning rather than their literal form. For example, "haunted" is translated as "계속 떠오른다" when it refers to "a thought or memory that repeatedly comes to mind", rather than as the literal ghost-related meaning "귀신이 나오는".

\paragraph{Honorific usage.}
Direct translation can produce repeated address forms in which a Korean name is followed by the polite suffix "씨" throughout a dialogue. Since such repetition can sound unnatural or overly explicit in Korean, we reduce it where appropriate, while preserving the relationship between speakers and the emotional tone of the original item.

\subsection{Ko-WinoGrande: Names and Everyday Objects}
\label{app:winogrande-rules}

English personal names are replaced with Korean given names (e.g., Jessica~$\rightarrow$~지희). The substitutions are kept
consistent within each item and checked to preserve the original coreference structure. Everyday objects that are uncommon or unnatural in Korean contexts are replaced with familiar Korean counterparts when the replacement does not change the commonsense relation required by the item (e.g., toaster oven~$\rightarrow$~전자레인지(microwave) as Koreans typically don't use toaster ovens). Sentence structure is also adjusted to avoid excessive repetition of names, which is more marked in Korean than in English.

\subsection{Ko-IFEval: Hybrid Adaptation and Korean-Specific Verifiers}
\label{app:ifeval-expansion}

\paragraph{Construction subsets.}
Ko-IFEval is constructed through a hybrid strategy. Items with indices 1,000-3,757 form the translated subset, where original IFEval prompts are adapted through the four-stage pipeline described in Section~\ref{3_2:adapt}. For these items, we translate and localize the prompt while preserving the original instruction-following constraint whenever it can be evaluated reliably in Korean. Localization includes replacing culturally unfamiliar genres with Korean counterparts. For example, an instruction asking for an English "limerick" is rewritten as a request for "5행시", a Korean five-line poem written from a given word or phrase. We also convert verifier arguments, such as comparison relations and target strings, so that the automatic checker can evaluate Korean outputs directly.

Items with indices 10,001-10,300 form the Korean-specific subset. For these items, authors write Korean prompts from scratch using public-domain Korean source material, including presidential speeches. We define additional instruction types that reflect Korean writing conventions and implement verifier code for checking these constraints automatically.

\paragraph{Instruction-type expansion.}
Table~\ref{tab:ifeval-groups} summarizes how the instruction-type inventory is expanded from 25 types in the original IFEval to 30 types in Ko-IFEval. The expansion is concentrated in instruction groups where direct transfer from English is insufficient, especially keyword constraints, length constraints, and detectable formatting. Other instruction groups are retained without structural change.

\begin{table*}[t]
\centering
\small
\begin{tabular}{p{3.6cm} c c p{8.5cm}}
\toprule
\textbf{Instruction group} & \textbf{IFEval} & \textbf{Ko-IFEval} & \textbf{Changes} \\
\midrule
Keywords & 4 & 6 &
Alphabet-based letter constraints are separated from Korean choseong(초성; leading consonant)-based constraints; a keyword-substitution check is added for Korean prompts. \\

Length Constraints & 4 & 6 &
Character-count checks are extended to distinguish whitespace-inclusive and whitespace-exclusive counting, which is necessary for Korean spacing conventions. \\

Detectable Format & 6 & 7 &
A multiple-choice formatting check is added. \\

Language & 1 & 1 &
No structural change; the required response language can be Korean or English. \\

Detectable Content & 2 & 2 &
No structural change. \\

Combination & 2 & 2 &
No structural change. \\

Change Cases & 3 & 3 &
No structural change. \\

Start with / End with & 2 & 2 &
No structural change. \\

Punctuation & 1 & 1 &
No structural change. \\
\midrule
\textbf{Total} & \textbf{25} & \textbf{30} &
+5 new or adapted instruction types. \\
\bottomrule
\end{tabular}
\caption{Instruction-type inventory in the original IFEval and Ko-IFEval. The "Changes" column summarizes Ko-IFEval-specific additions and adaptations.}
\label{tab:ifeval-groups}
\end{table*}

\paragraph{Verifier arguments.}
For Korean-specific instruction types, each item includes Korean-aware arguments in the \texttt{kwargs} field. For example, items with the \code{detectable\_format:number\_highlighted\_sections} instruction specify \texttt{num\_highlights} together with a Korean comparison relation such as "이상(greater than or equal to)" or "미만(less)". Length-constraint items also specify whether \texttt{num\_letters} should be counted with or without whitespace. The verifier code reads these fields directly, allowing the original IFEval scoring framework to be extended to Ko-IFEval with a small compatibility layer.

\section{Review Process and Quality Control}
\label{app:quality}

\subsection{Participants and Task Assignments}
Correction and review were conducted internally by members of the author team, rather than by external reviewers. Fifteen authors participated in dataset construction and review. Authors performing correction and cross-review were native Korean speakers with at least CEFR\footnote{Common European Framework of Reference for Languages.} B2-equivalent English proficiency. The team included members with Korean-English bilingual backgrounds, experience in English-speaking educational environments, and translation-related professional experience. Relevant backgrounds included Korean language education, mathematics education, computer science, statistics, physics, geography, psychology, economics, accounting, and history. Responsibilities were assigned with consideration of each subset's requirements and the participants' language and domain backgrounds.

\subsection{Review Procedure}
The correction and review procedures for Ko-ARC, Ko-GSM8K, Ko-EQ-Bench, and Ko-WinoGrande are described in Section~\ref{3_2:adapt}. Ko-LAMBADA followed a separate construction process based on Korean literary texts. Passages and answer candidates were manually reviewed for contextual support and grammatical compatibility. Ko-IFEval also followed a separate adaptation and verification process. Authors adapted 541 source prompts and wrote 300 additional Korean prompts, manually checking their consistency with instruction identifiers and verifier arguments and adapting the associated checks for Korean.

\subsection{Recorded Revision Statistics}
We quantify documented editing as the fraction of aligned items with a change to at least one task field between archived versions. Changes include localization, word edits, and answer corrections. Items are aligned by their original identifiers or source prompts. We normalize line endings, Unicode representation, and leading or trailing whitespace, and compare serialized answer choices by their contents. Table~\ref{tab:revision_stats} reports confirmed changes in records associated with initial editing, localization, and cross-review.

The high initial correction rates show that author revisions were needed across a large proportion of the machine-translated items. These revisions included restoring information omitted during translation, correcting inconsistent formatting, and revising wording that obscured the intended meaning. In particular, missing task-relevant information had to be recovered from the English source to preserve the conditions needed to answer the question. These examples illustrate the substantive role of author-led correction in preparing the translated drafts for Korean evaluation.

\begin{table*}[t]
\centering
\small
\begin{tabular}{lrrrr}
\toprule
Dataset & Aligned items & Correction
& Localization & Cross-review \\
\midrule
Ko-ARC-Easy
& 2,376
& 2,056 (86.5\%)
& 643 (27.1\%)
& 72 (3.0\%) \\

Ko-ARC-Challenge
& 1,172
& 720 (61.4\%)
& 73 (6.2\%)
& 42 (3.6\%) \\

Ko-GSM8K
& 1,319
& 1,291 (97.9\%)
& 830 (62.9\%)
& 358 (27.1\%) \\

Ko-EQ-Bench
& 171
& 171 (100.0\%)
& 165 (96.5\%)
& 14 (8.2\%) \\
\bottomrule
\end{tabular}
\caption{
Recorded item revisions at each construction stage,
reported as counts and percentages of aligned items.
Ko-ARC-Challenge statistics cover the 1,172-item review pool before five items were excluded from the final dataset. The separate Ko-IFEval process is not included.
}
\label{tab:revision_stats}
\end{table*}

\begin{table*}[t]
\centering
\small
\begin{tabular}{l l l l}
\toprule
\textbf{Provider} & \textbf{Model} & \textbf{Size} & \textbf{Hugging Face checkpoint} \\
\midrule
\multicolumn{4}{l}{\emph{Korean-developed LLMs}} \\
KT      & Mi:dm 2.0 Mini             & 2.3B     & \href{https://huggingface.co/K-intelligence/Midm-2.0-Mini-Instruct}{\texttt{K-intelligence/Midm-2.0-Mini-Instruct}} \\
      & Mi:dm 2.0 Base             & 11.5B    & \href{https://huggingface.co/K-intelligence/Midm-2.0-Base-Instruct}{\texttt{K-intelligence/Midm-2.0-Base-Instruct}} \\
LG      & EXAONE-4.0-1.2B            & 1.2B     & \href{https://huggingface.co/LGAI-EXAONE/EXAONE-4.0-1.2B}{\texttt{LGAI-EXAONE/EXAONE-4.0-1.2B}} \\
      & EXAONE-4.0-32B             & 32B      & \href{https://huggingface.co/LGAI-EXAONE/EXAONE-4.0-32B}{\texttt{LGAI-EXAONE/EXAONE-4.0-32B}} \\
Kakao   & Kanana-2-30b-a3b-base-2601              & 30B-A3B  & \href{https://huggingface.co/kakaocorp/kanana-2-30b-a3b-base-2601}{\texttt{kakaocorp/kanana-2-30b-a3b-base-2601}} \\
        & Kanana-2-30b-a3b-mid-2601               & 30B-A3B  & \href{https://huggingface.co/kakaocorp/kanana-2-30b-a3b-mid-2601}{\texttt{kakaocorp/kanana-2-30b-a3b-mid-2601}} \\
        & Kanana-2-30b-a3b-instruct-2601          & 30B-A3B  & \href{https://huggingface.co/kakaocorp/kanana-2-30b-a3b-instruct-2601}{\texttt{kakaocorp/kanana-2-30b-a3b-instruct-2601}} \\
Naver   & HyperCLOVAX-SEED-Text-Instruct-0.5B      & 0.5B     & \href{https://huggingface.co/naver-hyperclovax/HyperCLOVAX-SEED-Text-Instruct-0.5B}{\texttt{HyperCLOVAX-SEED-Text-Instruct-0.5B}} \\
   & HyperCLOVAX-SEED-Text-Instruct-1.5B      & 1.5B     & \href{https://huggingface.co/naver-hyperclovax/HyperCLOVAX-SEED-Text-Instruct-1.5B}{\texttt{HyperCLOVAX-SEED-Text-Instruct-1.5B}} \\
SKT     & A.X 4.0 Light              & 7B       & \href{https://huggingface.co/skt/A.X-4.0-Light}{\texttt{skt/A.X-4.0-Light}} \\
\midrule
\multicolumn{4}{l}{\emph{Multilingual LLMs}} \\
Qwen    & Qwen3.5-0.8B-Base         & 0.8B     & \href{https://huggingface.co/Qwen/Qwen3.5-0.8B-Base}{\texttt{Qwen/Qwen3.5-0.8B-Base}} \\
    & Qwen3.5-0.8B              & 0.8B     & \href{https://huggingface.co/Qwen/Qwen3.5-0.8B}{\texttt{Qwen/Qwen3.5-0.8B}} \\
    & Qwen3.5-2B-Base           & 2B       & \href{https://huggingface.co/Qwen/Qwen3.5-2B-Base}{\texttt{Qwen/Qwen3.5-2B-Base}} \\
    & Qwen3.5-2B                & 2B       & \href{https://huggingface.co/Qwen/Qwen3.5-2B}{\texttt{Qwen/Qwen3.5-2B}} \\
    & Qwen3.5-4B-Base           & 4B       & \href{https://huggingface.co/Qwen/Qwen3.5-4B-Base}{\texttt{Qwen/Qwen3.5-4B-Base}} \\
    & Qwen3.5-4B                & 4B       & \href{https://huggingface.co/Qwen/Qwen3.5-4B}{\texttt{Qwen/Qwen3.5-4B}} \\
    & Qwen3.5-9B-Base           & 9B       & \href{https://huggingface.co/Qwen/Qwen3.5-9B-Base}{\texttt{Qwen/Qwen3.5-9B-Base}} \\
    & Qwen3.5-9B                & 9B       & \href{https://huggingface.co/Qwen/Qwen3.5-9B}{\texttt{Qwen/Qwen3.5-9B}} \\
    & Qwen3.5-27B               & 27B      & \href{https://huggingface.co/Qwen/Qwen3.5-27B}{\texttt{Qwen/Qwen3.5-27B}} \\
    & Qwen3.5-35B-A3B-Base              & 35B-A3B  & \href{https://huggingface.co/Qwen/Qwen3.5-35B-A3B-Base}{\texttt{Qwen/Qwen3.5-35B-A3B-Base}} \\
    & Qwen3.5-35B-A3B                  & 35B-A3B  & \href{https://huggingface.co/Qwen/Qwen3.5-35B-A3B}{\texttt{Qwen/Qwen3.5-35B-A3B}} \\
Google   & Gemma-4-E2B                & 5B      & \href{https://huggingface.co/google/gemma-4-E2B}{\texttt{google/gemma-4-E2B}} \\
         & Gemma-4-E2B-it             & 5B      & \href{https://huggingface.co/google/gemma-4-E2B-it}{\texttt{google/gemma-4-E2B-it}} \\
         & Gemma-4-E4B                & 8B       & \href{https://huggingface.co/google/gemma-4-E4B}{\texttt{google/gemma-4-E4B}} \\
         & Gemma-4-E4B-it             & 8B       & \href{https://huggingface.co/google/gemma-4-E4B-it}{\texttt{google/gemma-4-E4B-it}} \\
         & Gemma-4-26B-A4B                   & 26B-A4B  & \href{https://huggingface.co/google/gemma-4-26B-A4B}{\texttt{google/gemma-4-26B-A4B}} \\
         & Gemma-4-26B-A4B-it                 & 26B-A4B  & \href{https://huggingface.co/google/gemma-4-26B-A4B-it}{\texttt{google/gemma-4-26B-A4B-it}} \\
         & Gemma-4-31B                & 31B      & \href{https://huggingface.co/google/gemma-4-31B}{\texttt{google/gemma-4-31B}} \\
         & Gemma-4-31B-it             & 31B      & \href{https://huggingface.co/google/gemma-4-31B-it}{\texttt{google/gemma-4-31B-it}} \\
\bottomrule
\end{tabular}
\caption{Open-weight models evaluated. Sizes are reported in billions of parameters. For mixture-of-experts models, we also include the active-parameter shorthand reported by the developer, such as \texttt{30B-A3B}. All checkpoints are accessed through the Hugging Face Hub.}
\label{tab:open-models}
\end{table*}

\section{Open-Weight Models Evaluated}
\label{app:models}

Table~\ref{tab:open-models} lists all open-weight models included in our evaluation, grouped by provider. We report the model size and the exact Hugging Face checkpoint used for evaluation. The Korean-developed group spans five model families released by Korean companies, while the multilingual group includes two widely used model families with multilingual capabilities.

\section{Detailed Evaluation Protocol}
\label{app:evaluation_protocol}

\paragraph{Common settings.}
All evaluations are conducted with \texttt{lm-evaluation-harness}. We use the test split for every task. Each reported result is based on a single run. For generation-based tasks, we use greedy decoding with \texttt{do\_sample=False} and \texttt{temperature=0.0}.

\paragraph{Models.}
Closed API models are evaluated through their official APIs. We use Claude Haiku 4.5 (\texttt{claude-haiku-4-5-20251001}) and Claude Opus 4.7 (\texttt{claude-opus-4-7}) through the Anthropic API, and GPT-5-mini (\texttt{gpt-5-mini-2025-08-07}) and GPT-5.5 (\texttt{gpt-5.5-2026-04-23}) through the OpenAI API. Open-weight models are evaluated using the Hugging Face Transformers backend of \texttt{lm-evaluation-harness}; each model is identified by its exact Hugging Face checkpoint. Open-weight model inference uses bfloat16 precision on a single NVIDIA B200 GPU with 180 GB memory.

\begin{table*}[t]
\centering
\small
\begin{tabular}{l c l c p{5.4cm}}
\toprule
\textbf{Task} & \textbf{Few-shot} & \textbf{Output type} &
\textbf{Max gen.\ toks} & \textbf{Metric} \\
\midrule
Ko-ARC-Easy & 5 & multiple choice & -- &
\texttt{acc} / \texttt{acc\_norm} / \texttt{acc\_npsq} \\
Ko-ARC-Challenge & 5 & multiple choice & -- &
\texttt{acc} / \texttt{acc\_norm} / \texttt{acc\_npsq} \\
Ko-WinoGrande & 0 & multiple choice & -- &
\texttt{acc} / \texttt{acc\_norm} / \texttt{acc\_npsq} \\
Ko-LAMBADA & 0 & multiple choice & -- &
\texttt{acc} / \texttt{acc\_norm} / \texttt{acc\_npsq} \\
Ko-GSM8K & 5 & generate-until & 2048 &
\texttt{strict-match} / \texttt{flexible-extract} \\
Ko-IFEval & 0 & generate-until & 1280 &
prompt-level and instruction-level strict/loose accuracy \\
Ko-EQ-Bench & 0 & generate-until & 80 / 1024 &
\texttt{eqbench} / \texttt{percent\_parseable} \\
Ko-ARC-gen & 5 & generate-until & 512 &
\texttt{exact\_match} \\
Ko-WinoGrande-gen & 0 & generate-until & 512 &
\texttt{exact\_match} \\
Ko-LAMBADA-gen & 0 & generate-until & 512 &
\texttt{exact\_match} \\
\bottomrule
\end{tabular}
\caption{Detailed task configuration. Original multiple-choice tasks are evaluated only for open-weight models with log-likelihood access. Generative multiple-choice variants are used for both open-weight and closed API models. For Ko-EQ-Bench, the maximum generation length is 80 tokens for open-weight models and 1024 tokens for closed API models.}
\label{tab:detailed_eval_tasks}
\end{table*}
\paragraph{Stop sequences.}
For the generative variants of Ko-ARC, Ko-WinoGrande, and Ko-LAMBADA, generation stops at one of \texttt{\textbackslash n}, \texttt{</s>}, or \texttt{<|im\_end|>}. For Ko-GSM8K, generation stops at \texttt{문제:}, \texttt{</s>}, or \texttt{<|im\_end|>}. Ko-IFEval uses no explicit stop sequence. Ko-EQ-Bench stops at a blank line
(\texttt{\textbackslash n\textbackslash n}), with maximum generation lengths of 80 tokens for open-weight models and 1024 tokens for closed API models. 


\paragraph{Prompt formatting conditions.}
For open-weight models, we distinguish between plain prompting and chat-template prompting. Plain prompting is used for all open-weight checkpoints. For instruction-tuned or chat-oriented checkpoints with a defined chat template, we additionally evaluate chat-template prompting using the \texttt{--apply\_chat\_template} option in \texttt{lm-evaluation-harness}. This option formats each benchmark prompt according to the checkpoint-specific chat template by adding model-specific role markers, special tokens, or generation prefixes. When few-shot examples are used, the demonstrations and the test instance are represented as alternating user and model turns.

For example, for a Gemma instruction-tuned checkpoint, a task prompt containing Korean benchmark-specific fields such as \texttt{질문:} and \texttt{답변:} is wrapped as:
\begin{quote}
\small
\texttt{<bos><|turn>user}\\
\texttt{[benchmark prompt]}\\
\texttt{<turn|>}\\
\texttt{<|turn>model}
\end{quote}
Here, the Korean field labels are part of the benchmark prompt, whereas the turn markers and special tokens are added by the chat template.

Closed API models are evaluated only through their official chat-style APIs. Since these APIs do not expose tokenizer-level chat-template control or a comparable plain prompting interface, we cannot evaluate closed models under the plain prompting condition. Closed-model results should therefore be interpreted as chat-style prompting results, and compared most directly with the open-weight chat-template condition rather than the open-weight plain condition.

We report additional results for six instruction-tuned Qwen3.5 models with \texttt{enable\_thinking=False} on Ko-GSM8K, Ko-EQ-Bench, and the four generative MCQA tasks, while retaining chat templates. These results are reported separately from the default-chat results.

\paragraph{Prompting and answer extraction.}
For original multiple-choice tasks, open-weight models are evaluated using log-likelihood-based scoring. For Ko-ARC, the model receives the question prompt, and each answer choice is scored as a candidate continuation. For Ko-WinoGrande and Ko-LAMBADA, each candidate is inserted into the blank and scored by log-likelihood comparison. 

We evaluate generative multiple-choice variants for both open-weight and closed API models.
Ko-ARC-Easy-gen and Ko-ARC-Challenge-gen (generation versions) present four labeled options and require the model to output a single label:

\begin{quote}
\small
\begin{verbatim}
질문: {question}
A. {choices.text[0]}
B. {choices.text[1]}
C. {choices.text[2]}
D. {choices.text[3]}
반드시 A, B, C, D 중 하나의 문자로만 답하세요.
정답:
\end{verbatim}
\end{quote}

The final answer is extracted with the following regular expression:

\begin{quote}
\small
\verb!(?i)(?:정답\s*[:：]?\s*)?([A-D])(?:\s*(?:번|입니다|이에요|가)|[\.)\]:：,\s]|$)!
\end{quote}

Ko-WinoGrande-gen presents two labeled options and requires the model to output either A or B:

\begin{quote}
\small
\begin{verbatim}
문장의 빈칸에 들어갈 말로 더 자연스러운 선택지를 고르세요.
문장: {sentence}
A. {option1}
B. {option2}
반드시 A 또는 B 중 하나의 문자로만 답하세요.
정답:
\end{verbatim}
\end{quote}

Ko-LAMBADA-gen uses the same two-choice format, with the masked passage
provided as the sentence field:

\begin{quote}
\small
\begin{verbatim}
문장의 빈칸에 들어갈 말로 더 자연스러운 선택지를 고르세요.
문장: {text}
A. {option_A}
B. {option_B}
반드시 A 또는 B 중 하나의 문자로만 답하세요.
정답:
\end{verbatim}
\end{quote}

For Ko-WinoGrande-gen and Ko-LAMBADA-gen, the final answer is extracted with the same regular expression, except that the valid label range is A-B:

\begin{quote}
\small
\verb!(?i)(?:정답\s*[:：]?\s*)?([A-B])(?:\s*(?:번|입니다|이에요|가)|[\.)\]:：,\s]|$)!
\end{quote}

If no valid label is matched, the response is counted as incorrect.

For Ko-GSM8K, the prompt format is \texttt{문제: \{question\}\textbackslash n답:}. 
We use two independent answer extractors: a strict extractor that requires an explicit final-answer marker and a flexible extractor that selects the final numeric match regardless of whether that marker is present. During answer normalization, commas, the Korean currency marker "원", text before \texttt{\#\#\#\#}, and a trailing period are ignored.

Ko-IFEval uses the \texttt{prompt} field directly and is evaluated with deterministic instruction-following verifiers. 
Ko-EQ-Bench also uses the \texttt{prompt} field directly. The model is expected to output four emotion-score pairs, one per line, in the form \texttt{emotion: score}. 
The reported EQ-Bench scores are reproduced by a line-based parser that allows spaces within Korean emotion labels and whitespace before the colon. An output is parseable when the extracted label set consists of exactly the four reference emotions; otherwise, both \texttt{eqbench} and \texttt{percent\_parseable} are set to zero.

\paragraph{Metric definitions.}
For multiple-choice tasks, \texttt{acc} denotes the fraction of examples for which the correct option has the highest log-likelihood. \texttt{acc\_norm} applies length normalization to the per-choice log-likelihoods before comparison. \texttt{acc\_npsq} uses Normalized Probability Shift by the Question (NPSQ; \citealp{cho2026choices}). For a question-related input $Q$, choice-related input $C$, and answer candidate $x$, NPSQ is defined as

{\small
\[
\mathrm{NPSQ}(Q,C,x)
=
\frac{\log P(x \mid Q,C) - \log P(x \mid C)}
{-\log P(x \mid C)}.
\]
}

The prediction is the option with the highest NPSQ score. This scoring rule measures the relative increase in the model's preference for each choice after the question is included, compared with a question-free baseline that scores the choices without the question. It therefore separates question-driven evidence from choice-only preferences such as choice-length bias or surface-form familiarity.

For Ko-GSM8K, we report two exact-match variants: \texttt{strict-match} and \texttt{flexible-extract}. The \texttt{strict-match} metric counts an example as correct only when the response contains an explicit final-answer marker of the form \texttt{\#\#\#\# number} and the extracted number matches the gold answer after normalization. 
The \texttt{flexible-extract} metric independently extracts the final numeric match in the response, regardless of whether an explicit final-answer marker is present, and compares it with the gold answer after normalization.

For Ko-IFEval, we report four instruction-following accuracies. Prompt-level strict accuracy counts a prompt as correct only when all verifiable instructions in that prompt are satisfied under the strict verifier. Instruction-level strict accuracy is the fraction of individual verifiable instructions that are satisfied. Prompt-level loose accuracy and instruction-level loose accuracy use the same two aggregation levels, but apply the loose verifier, which allows common surface variations in the response before checking whether each instruction is followed.

For Ko-EQ-Bench, we report two metrics: \texttt{eqbench} and \texttt{percent\_parseable}. The model is instructed to output four emotion-score pairs in the form \texttt{emotion: score}, with integer scores on a 0-10 scale. 
\texttt{percent\_parseable} is the percentage of examples for which the extracted label set consists of exactly the four reference emotions.
For parseable outputs, \texttt{eqbench} measures how closely the predicted emotion intensity scores match the reference scores. For each matched emotion, we compute the absolute difference between the predicted and reference scores and apply the same scaled penalty function used by the EQ-Bench scorer. The example-level score is 100 for exact agreement and can be negative for sufficiently large errors. Unparseable outputs receive \texttt{eqbench}=0 and \texttt{percent\_parseable}=0 before aggregation, so the reported \texttt{eqbench} score reflects both emotion-score accuracy and output format compliance.

\section{Qwen3.5 Behavior under Chat Prompting}
\label{app:qwen-traces}

We inspect representative Ko-GSM8K outputs from Qwen3.5-9B and Qwen3.5-0.8B under chat prompting. The example uses a simple arithmetic problem about selling duck eggs. Both models have enough information to solve the problem, and the correct answer is 18,000. However, their generation behavior differs. Qwen3.5-0.8B directly solves the problem in Korean and reaches the required final answer format. In contrast, Qwen3.5-9B begins with a long English reasoning trace and does not produce the final answer in the recorded output. This example illustrates a failure to produce the final answer under default chat prompting; the thinking-off results in Table~\ref{tab:qwen_thinking} show that performance depends on the evaluated generation settings.

\paragraph{Qwen3.5-0.8B.}
Qwen3.5-0.8B produces a short Korean solution and reaches the final answer.

\begin{quote}
\small
보리의 하루는 16개의 알을 낳습니다. 아침 식사로 3개, 머핀으로 4개, 나머지는 직거래 장터에서 16 - 3 - 4 = 9개의 알을 판매합니다.

매일 직거래 장터에서 벌이는 돈은 9개 x 2000원 = <<9*2000=18000>>18,000원입니다.

\#\#\#\# 18000
\end{quote}

\paragraph{Qwen3.5-9B.}
Qwen3.5-9B begins solving the same problem in English and produces a long reasoning trace. In the recorded output, the model does not reach the final answer.

\begin{quote}
\small
Thinking Process:

1. Analyze the Request: The user is presenting a math word problem in Korean and expects a solution in a specific format ...

2. Analyze the Problem:\\
* Duck lays 16 eggs per day.\\
* Duck eats 3 eggs for breakfast.\\
* Duck makes muffins with 4 eggs for friends.\\
* Remaining eggs are sold at the direct market.\\
* Selling price: 2000 won per egg.

3. Step-by-Step Calculation:\\
* Total eggs laid: 16\\
* Eggs eaten: 3\\
* Eggs for muffins: 4\\
* Eggs remaining for sale: $16 - 3 - 4$\\
* Calculation: $16 - 3 = 13$, $13 - 4 =$ 
\end{quote}

\section{Full Experimental Results}
\label{app:full-results}


This section reports per-model \kbs results and model-group summaries for language comparisons.
Tables~\ref{tab:mcqa_plain} and~\ref{tab:mcqa_chat} report open-weight log-likelihood MCQA results under plain and default chat prompting, respectively.
Tables~\ref{tab:generation_plain} and~\ref{tab:generation_chat} report open-weight generation results under the two prompt formats.
Table~\ref{tab:api_total} reports closed API results for generative MCQA and generation tasks.
Table~\ref{tab:generative_mcqa_chat} reports open-weight generative MCQA results under default chat prompting. Table~\ref{tab:qwen_nothink} reports Qwen3.5 thinking-off results for both generative MCQA and generation tasks. Table~\ref{tab:language_comparison_appendix} summarizes language comparisons for ARC, GSM8K, WinoGrande, and EQ-Bench.

The thinking-off comparison in Table~\ref{tab:qwen_nothink} focuses on Ko-GSM8K, Ko-EQ-Bench, and generative MCQA, where the four larger Qwen3.5 models obtain near-zero scores under default chat prompting (Tables~\ref{tab:generation_chat} and~\ref{tab:generative_mcqa_chat}). Ko-IFEval does not exhibit the same near-zero pattern and is not included in this comparison.

\begin{table*}[htbp]
\centering
\resizebox{\textwidth}{!}{%
\begin{tabular}{@{}lrrrrrrrrrrrrr@{}}
\toprule
\multicolumn{1}{c}{} & \multicolumn{1}{c}{} & \multicolumn{3}{c}{\textbf{Ko-WinoGrande}} & \multicolumn{3}{c}{\textbf{Ko-LAMBADA}} & \multicolumn{3}{c}{\textbf{Ko-ARC-Easy}} & \multicolumn{3}{c}{\textbf{Ko-ARC-Challenge}} \\ \cmidrule(l){3-5}\cmidrule(l){6-8}\cmidrule(l){9-11}\cmidrule(l){12-14}
\multicolumn{1}{c}{\multirow{-2}{*}{\textbf{Model}}} & \multicolumn{1}{c}{\multirow{-2}{*}{\textbf{\begin{tabular}[c]{@{}c@{}}Parameter \\ size\end{tabular}}}} & \multicolumn{1}{c}{\textbf{acc}} & \multicolumn{1}{c}{\textbf{\begin{tabular}[c]{@{}c@{}}acc\\ \_norm\end{tabular}}} & \multicolumn{1}{c}{\textbf{\begin{tabular}[c]{@{}c@{}}acc\\ \_npsq\end{tabular}}} & \multicolumn{1}{c}{\textbf{acc}} & \multicolumn{1}{c}{\textbf{\begin{tabular}[c]{@{}c@{}}acc\\ \_norm\end{tabular}}} & \multicolumn{1}{c}{\textbf{\begin{tabular}[c]{@{}c@{}}acc\\ \_npsq\end{tabular}}} & \multicolumn{1}{c}{\textbf{acc}} & \multicolumn{1}{c}{\textbf{\begin{tabular}[c]{@{}c@{}}acc\\ \_norm\end{tabular}}} & \multicolumn{1}{c}{\textbf{\begin{tabular}[c]{@{}c@{}}acc\\ \_npsq\end{tabular}}} & \multicolumn{1}{c}{\textbf{acc}} & \multicolumn{1}{c}{\textbf{\begin{tabular}[c]{@{}c@{}}acc\\ \_norm\end{tabular}}} & \multicolumn{1}{c}{\textbf{\begin{tabular}[c]{@{}c@{}}acc\\ \_npsq\end{tabular}}} \\ \midrule\midrule
\begin{tabular}[c]{@{}l@{}}HyperCLOVAX-SEED-Text-\\ Instruct-0.5B\end{tabular} & 0.5B & 55.72 & 55.88 & 54.62 & 93.30 & 92.95 & 87.72 & 58.71 & 58.67 & 61.66 & 32.39 & 38.73 & 38.05 \\
\begin{tabular}[c]{@{}l@{}}HyperCLOVAX-SEED-Text-\\ Instruct-1.5B\end{tabular} & 1.5B & 56.83 & 57.06 & 54.70 & 90.02 & 89.22 & 84.17 & 60.35 & 61.45 & 59.51 & 34.53 & 40.62 & 38.13 \\ \midrule
A.X 4.0 Light & 7B & 64.96 & 63.38 & 62.75 & 96.59 & 96.32 & 93.13 & 78.41 & 78.87 & 79.46 & 55.87 & 59.73 & 61.27 \\ \midrule
Mi:dm 2.0 Mini & 2.3B & 61.25 & 60.62 & 58.80 & 96.76 & 96.54 & 93.48 & 75.55 & 76.60 & 78.11 & 48.33 & 54.84 & 58.70 \\
Mi:dm 2.0 Base & 11.5B & 66.22 & 65.04 & 63.69 & 98.32 & 98.18 & 96.72 & 81.57 & 82.66 & 85.23 & 58.78 & 64.18 & 65.55 \\ \midrule
EXAONE-4.0-1.2B & 1.2B & 51.93 & 51.78 & 51.86 & 74.68 & 74.01 & 67.58 & 36.79 & 36.79 & 40.07 & 22.62 & 28.02 & 25.79 \\
EXAONE-4.0-32B & 32B & 60.06 & 59.83 & 58.64 & 94.41 & 93.93 & 90.33 & 78.33 & 80.01 & 81.10 & 55.10 & 59.55 & 62.90 \\ \midrule
Kanana-2-30b-a3b-base-2601 & 30B-A3B & 67.88 & 67.48 & 66.77 & {\color[HTML]{FE0000} 98.76} & {\color[HTML]{FE0000} 98.67} & 97.78 & 78.37 & 79.00 & 81.90 & 54.33 & 60.93 & 63.15 \\
Kanana-2-30b-a3b-mid-2601 & 30B-A3B & 69.06 & 67.88 & 65.67 & 98.67 & 98.58 & {\color[HTML]{FE0000} 97.87} & 79.42 & 78.75 & 82.11 & 54.41 & 60.50 & 63.50 \\
Kanana-2-30b-a3b-instruct-2601 & 30B-A3B & 71.27 & 70.40 & 68.90 & 98.45 & 98.36 & 97.47 & 80.05 & 80.30 & 81.36 & 58.61 & 63.24 & 63.41 \\ \midrule
Qwen3.5-0.8B-Base & 0.8B & 53.35 & 53.99 & 50.91 & 89.36 & 88.38 & 82.40 & 51.14 & 51.98 & 56.27 & 32.99 & 36.68 & 37.02 \\
Qwen3.5-0.8B & 0.8B & 55.01 & 55.17 & 53.43 & 87.94 & 86.79 & 79.82 & 47.43 & 49.41 & 52.82 & 29.14 & 33.16 & 32.73 \\
Qwen3.5-2B-Base & 2B & 57.93 & 57.06 & 52.88 & 93.61 & 93.08 & 86.79 & 60.35 & 62.46 & 66.12 & 38.48 & 44.47 & 44.56 \\
Qwen3.5-2B & 2B & 57.77 & 57.46 & 55.09 & 92.68 & 91.89 & 86.34 & 55.98 & 58.38 & 61.62 & 33.16 & 38.65 & 39.59 \\
Qwen3.5-4B-Base & 4B & 62.35 & 61.01 & 59.35 & 96.14 & 95.70 & 90.82 & 73.36 & 74.54 & 78.16 & 49.02 & 56.30 & 56.81 \\
Qwen3.5-4B & 4B & 61.41 & 61.25 & 58.33 & 95.34 & 94.95 & 90.78 & 72.05 & 73.36 & 77.10 & 47.90 & 55.10 & 56.21 \\
Qwen3.5-9B-Base & 9B & 65.04 & 63.85 & 59.67 & 96.45 & 96.50 & 91.57 & 77.74 & 79.63 & 81.94 & 54.84 & 61.78 & 63.84 \\
Qwen3.5-9B & 9B & 64.64 & 63.85 & 61.64 & 96.67 & 96.36 & 93.35 & 77.10 & 78.75 & 82.24 & 53.64 & 61.10 & 63.58 \\
Qwen3.5-27B & 27B & 71.51 & 70.64 & 68.59 & 97.92 & 97.83 & 95.88 & 80.64 & 83.33 & 84.89 & 58.36 & 64.61 & 67.27 \\
Qwen3.5-35B-A3B-Base & 35B-A3B & 69.30 & 67.72 & 64.48 & 97.38 & 97.38 & 95.30 & 81.19 & 83.50 & 84.98 & 60.24 & 66.67 & 70.01 \\
Qwen3.5-35B-A3B & 35B-A3B & 67.17 & 66.93 & 65.83 & 97.47 & 97.25 & 95.26 & 80.60 & 82.87 & 84.43 & 58.53 & 66.84 & 68.04 \\ \midrule
Gemma-4-E2B & 5B & 59.91 & 58.80 & 58.25 & 96.05 & 95.39 & 91.66 & 65.78 & 66.84 & 69.19 & 40.96 & 46.96 & 48.67 \\
Gemma-4-E2B-it & 5B & 51.22 & 50.99 & 50.67 & 70.20 & 70.02 & 60.98 & 39.94 & 40.70 & 40.91 & 26.39 & 30.51 & 29.22 \\
Gemma-4-E4B & 8B & 66.30 & 65.75 & 63.85 & 97.65 & 97.47 & 94.32 & 73.95 & 75.55 & 78.24 & 50.21 & 55.18 & 59.98 \\
Gemma-4-E4B-it & 8B & 51.78 & 51.14 & 49.65 & 76.19 & 75.26 & 64.26 & 49.24 & 49.87 & 50.51 & 27.42 & 34.70 & 33.42 \\
Gemma-4-26B-A4B & 26B-A4B & 70.56 & 68.98 & 67.64 & 98.36 & 98.09 & 96.67 & 80.93 & 82.24 & 86.03 & 56.98 & 64.01 & 67.10 \\
Gemma-4-26B-A4B-it & 26B-A4B & 51.14 & 51.54 & 49.49 & 82.31 & 82.13 & 67.36 & 56.78 & 57.16 & 58.29 & 35.99 & 40.96 & 41.56 \\
Gemma-4-31B & 31B & {\color[HTML]{FE0000} 74.74} & {\color[HTML]{FE0000} 73.17} & {\color[HTML]{FE0000} 71.98} & 98.63 & 98.63 & 97.07 & {\color[HTML]{FE0000} 83.54} & {\color[HTML]{FE0000} 85.94} & {\color[HTML]{FE0000} 88.13} & {\color[HTML]{FE0000} 61.18} & {\color[HTML]{FE0000} 67.78} & {\color[HTML]{FE0000} 70.61} \\
Gemma-4-31B-it & 31B & 52.64 & 52.09 & 52.49 & 81.91 & 82.00 & 67.36 & 51.26 & 52.53 & 52.78 & 32.82 & 39.67 & 39.59 \\ \midrule\midrule
\multicolumn{2}{c}{\textbf{Average}} & 61.69 & 61.06 & 59.33 & 92.49 & 92.13 & 87.39 & 67.81 & 69.04 & 71.21 & 45.63 & 51.57 & 52.77 \\ \midrule
\multicolumn{2}{c}{\textbf{Average (chat-template subset)}} & 59.59 & 59.17 & 57.73 & 90.06 & 89.67 & 84.00 & 64.49 & 65.65 & 67.34 & 42.75 & 48.57 & 49.17 \\ \bottomrule
\end{tabular}%
}
\caption{Open-weight multiple-choice results under plain prompting. The chat-template subset average uses the same model subset as the chat-template prompting results in Table~\ref{tab:mcqa_chat}. The highest score in each metric column is highlighted in red.}
\label{tab:mcqa_plain}
\end{table*}

\begin{table*}[htbp]
\centering
\resizebox{\textwidth}{!}{%
\begin{tabular}{@{}lrrrrrrrrrrrrr@{}}
\toprule
\multicolumn{1}{c}{} & \multicolumn{1}{c}{} & \multicolumn{3}{c}{\textbf{Ko-WinoGrande}} & \multicolumn{3}{c}{\textbf{Ko-LAMBADA}} & \multicolumn{3}{c}{\textbf{Ko-ARC-Easy}} & \multicolumn{3}{c}{\textbf{Ko-ARC-Challenge}} \\ \cmidrule(l){3-5}\cmidrule(l){6-8}\cmidrule(l){9-11}\cmidrule(l){12-14}
\multicolumn{1}{c}{\multirow{-2}{*}{\textbf{Model}}} & \multicolumn{1}{c}{\multirow{-2}{*}{\textbf{\begin{tabular}[c]{@{}c@{}}Parameter\\ size\end{tabular}}}} & \multicolumn{1}{c}{\textbf{acc}} & \multicolumn{1}{c}{\textbf{\begin{tabular}[c]{@{}c@{}}acc\\ \_norm\end{tabular}}} & \multicolumn{1}{c}{\textbf{\begin{tabular}[c]{@{}c@{}}acc\\ \_npsq\end{tabular}}} & \multicolumn{1}{c}{\textbf{acc}} & \multicolumn{1}{c}{\textbf{\begin{tabular}[c]{@{}c@{}}acc\\ \_norm\end{tabular}}} & \multicolumn{1}{c}{\textbf{\begin{tabular}[c]{@{}c@{}}acc\\ \_npsq\end{tabular}}} & \multicolumn{1}{c}{\textbf{acc}} & \multicolumn{1}{c}{\textbf{\begin{tabular}[c]{@{}c@{}}acc\\ \_norm\end{tabular}}} & \multicolumn{1}{c}{\textbf{\begin{tabular}[c]{@{}c@{}}acc\\ \_npsq\end{tabular}}} & \multicolumn{1}{c}{\textbf{acc}} & \multicolumn{1}{c}{\textbf{\begin{tabular}[c]{@{}c@{}}acc\\ \_norm\end{tabular}}} & \multicolumn{1}{c}{\textbf{\begin{tabular}[c]{@{}c@{}}acc\\ \_npsq\end{tabular}}} \\ \midrule\midrule
\begin{tabular}[c]{@{}l@{}}HyperCLOVAX-SEED-Text\\ -Instruct-0.5B\end{tabular} & 0.5B & 55.49 & 55.09 & 54.85 & 83.99 & 83.19 & 74.99 & 55.56 & 55.60 & 57.74 & 31.19 & 36.59 & 35.48 \\
\begin{tabular}[c]{@{}l@{}}HyperCLOVAX-SEED-Text\\ -Instruct-1.5B\end{tabular} & 1.5B & 57.46 & 56.35 & 55.09 & 83.86 & 83.55 & 78.76 & 57.91 & 59.72 & 58.88 & 33.25 & 40.19 & 37.45 \\ \midrule
A.X 4.0 Light & 7B & 58.72 & 59.12 & 57.77 & 86.56 & 85.81 & 78.49 & 75.46 & 77.15 & 77.36 & 50.64 & 56.56 & 56.47 \\ \midrule
Mi:dm 2.0 Mini & 2.3B & 59.67 & 58.49 & 54.78 & 83.55 & 81.82 & 66.83 & 72.39 & 73.70 & 75.25 & 45.59 & 52.19 & 51.76 \\
Mi:dm 2.0 Base & 11.5B & 64.33 & 63.06 & 60.62 & 86.43 & 85.06 & 68.65 & 77.53 & 74.45 & 79.59 & 54.67 & 58.61 & 61.70 \\ \midrule
EXAONE-4.0-1.2B & 1.2B & 50.28 & 50.20 & 51.22 & 76.28 & 73.84 & 62.48 & 58.12 & 49.41 & 53.54 & 34.19 & 41.30 & 38.73 \\
EXAONE-4.0-32B & 32B & 56.12 & 57.70 & 57.14 & 88.60 & 88.12 & {\color[HTML]{FE0000} 83.68} & 75.04 & 74.54 & 77.10 & 51.76 & 56.04 & 57.41 \\ \midrule
Kanana-2-30b-a3b-instruct-2601 & 30B-A3B & 61.09 & 61.25 & 60.46 & {\color[HTML]{FE0000} 89.00} & {\color[HTML]{FE0000} 88.60} & 82.79 & 76.39 & 74.03 & 75.25 & 54.67 & 58.61 & 58.10 \\ \midrule
Qwen3.5-0.8B & 0.8B & 51.78 & 51.70 & 50.83 & 73.66 & 71.80 & 61.20 & 45.88 & 47.18 & 48.49 & 29.05 & 34.96 & 33.59 \\
Qwen3.5-2B & 2B & 55.41 & 55.01 & 52.80 & 76.63 & 74.63 & 63.77 & 54.04 & 54.25 & 57.03 & 33.16 & 40.10 & 38.48 \\
Qwen3.5-4B & 4B & 56.83 & 55.41 & 53.04 & 76.67 & 75.26 & 61.73 & 47.98 & 41.16 & 44.15 & 36.50 & 43.10 & 43.70 \\
Qwen3.5-9B & 9B & 57.30 & 56.04 & 55.96 & 77.56 & 76.23 & 61.38 & 51.14 & 43.27 & 46.30 & 38.73 & 46.02 & 44.73 \\
Qwen3.5-27B & 27B & 59.35 & 58.33 & 57.77 & 82.97 & 81.29 & 70.51 & 55.77 & 47.90 & 54.55 & 47.13 & 51.59 & 52.70 \\
Qwen3.5-35B-A3B & 35B-A3B & 60.22 & 59.91 & 56.83 & 79.51 & 77.78 & 66.12 & 53.87 & 44.74 & 45.67 & 43.79 & 46.70 & 48.50 \\ \midrule
Gemma-4-E2B-it & 5B & 54.14 & 53.59 & 51.70 & 74.37 & 73.57 & 62.44 & 67.30 & 65.78 & 67.26 & 44.90 & 49.96 & 47.73 \\
Gemma-4-E4B-it & 8B & 59.12 & 58.96 & 58.09 & 80.04 & 79.38 & 68.60 & 75.63 & 75.00 & 76.98 & 52.79 & 59.13 & 56.90 \\
Gemma-4-26B-A4B-it & 26B-A4B & 61.88 & 62.19 & 52.96 & 79.25 & 79.16 & 61.55 & 77.61 & 78.96 & 79.13 & 57.58 & 63.58 & 62.81 \\
Gemma-4-31B-it & 31B & {\color[HTML]{FE0000} 66.22} & {\color[HTML]{FE0000} 65.19} & {\color[HTML]{FE0000} 61.17} & 83.10 & 82.00 & 71.62 & {\color[HTML]{FE0000} 80.35} & {\color[HTML]{FE0000} 79.92} & {\color[HTML]{FE0000} 80.51} & {\color[HTML]{FE0000} 62.04} & {\color[HTML]{FE0000} 65.30} & {\color[HTML]{FE0000} 65.12} \\ \midrule\midrule
\multicolumn{2}{c}{\textbf{Average}} & 58.08 & 57.64 & 55.73 & 81.22 & 80.06 & 69.20 & 64.33 & 62.04 & 64.15 & 44.53 & 50.03 & 49.52 \\ \bottomrule
\end{tabular}%
}
\caption{Open-weight multiple-choice results under chat-template prompting. The highest score in each metric column is highlighted in red.}
\label{tab:mcqa_chat}
\end{table*}
\begin{table*}[htbp]
\centering
\resizebox{\textwidth}{!}{%
\begin{tabular}{@{}lrrrrrrrrr@{}}
\toprule
\multicolumn{1}{c}{} & \multicolumn{1}{c}{} & \multicolumn{2}{c}{\textbf{Ko-GSM8K}} & \multicolumn{4}{c}{\textbf{Ko-IFEval}} & \multicolumn{2}{c}{\textbf{Ko-EQ-Bench}} \\ \cmidrule(l){3-4}\cmidrule(l){5-8}\cmidrule(l){9-10}
\multicolumn{1}{c}{\multirow{-2}{*}{\textbf{Model}}} & \multicolumn{1}{c}{\multirow{-2}{*}{\textbf{\begin{tabular}[c]{@{}c@{}}Parameter \\ size\end{tabular}}}} & \multicolumn{1}{c}{\textbf{\begin{tabular}[c]{@{}c@{}}strict\\ -match\end{tabular}}} & \multicolumn{1}{c}{\textbf{\begin{tabular}[c]{@{}c@{}}flexible\\ -extract\end{tabular}}} & \multicolumn{1}{c}{\textbf{\begin{tabular}[c]{@{}c@{}}prompt\\ \_level\\ \_strict\\ \_acc\end{tabular}}} & \multicolumn{1}{c}{\textbf{\begin{tabular}[c]{@{}c@{}}inst\\ \_level\\ \_strict\\ \_acc\end{tabular}}} & \multicolumn{1}{c}{\textbf{\begin{tabular}[c]{@{}c@{}}prompt\\ \_level\\ \_loose\\ \_acc\end{tabular}}} & \multicolumn{1}{c}{\textbf{\begin{tabular}[c]{@{}c@{}}inst\\ \_level\\ \_loose\\ \_acc\end{tabular}}} & \multicolumn{1}{c}{\textbf{eqbench}} & \multicolumn{1}{c}{\textbf{\begin{tabular}[c]{@{}c@{}}percent\\ \_parseable\end{tabular}}} \\ \midrule \midrule
HyperCLOVAX-SEED-Text-Instruct-0.5B & 0.5B & 4.09 & 3.79 & 12.49 & 25.99 & 14.63 & 29.06 & 7.90 & 89.47 \\
HyperCLOVAX-SEED-Text-Instruct-1.5B & 1.5B & 9.40 & 7.88 & 21.64 & 36.72 & 23.54 & 39.44 & 38.99 & 92.98 \\ \midrule
A.X 4.0 Light & 7B & 78.85 & 78.92 & {\color[HTML]{FE0000} 60.29} & {\color[HTML]{FE0000} 74.36} & {\color[HTML]{FE0000} 65.16} & {\color[HTML]{FE0000} 77.56} & 54.53 & {\color[HTML]{FE0000} 100.00} \\ \midrule
Mi:dm 2.0 Mini & 2.3B & 57.85 & 61.71 & 19.62 & 29.13 & 24.26 & 32.96 & 11.93 & 38.01 \\
Mi:dm 2.0 Base & 11.5B & 81.73 & 81.80 & 28.30 & 40.70 & 31.87 & 43.83 & 51.01 & 94.15 \\ \midrule
EXAONE-4.0-1.2B & 1.2B & 4.02 & 7.13 & 2.26 & 5.16 & 2.38 & 5.23 & -0.11 & 4.68 \\
EXAONE-4.0-32B & 32B & 78.24 & 79.38 & 37.22 & 52.13 & 40.55 & 54.77 & 58.88 & 94.15 \\ \midrule
Kanana-2-30b-a3b-base-2601 & 30B-A3B & 69.67 & 69.83 & 13.44 & 22.44 & 15.22 & 24.53 & 36.41 & {\color[HTML]{FE0000} 100.00} \\
Kanana-2-30b-a3b-mid-2601 & 30B-A3B & 76.72 & 76.65 & 17.60 & 33.31 & 19.74 & 35.82 & 30.54 & 65.50 \\
Kanana-2-30b-a3b-instruct-2601 & 30B-A3B & {\color[HTML]{FE0000} 83.02} & {\color[HTML]{FE0000} 83.17} & 30.44 & 49.48 & 36.15 & 54.08 & 55.30 & 98.83 \\ \midrule
Qwen3.5-0.8B-Base & 0.8B & 19.79 & 20.32 & 17.12 & 30.24 & 17.95 & 31.43 & -15.33 & 84.21 \\
Qwen3.5-0.8B & 0.8B & 18.80 & 19.11 & 22.95 & 39.93 & 26.99 & 43.90 & -0.35 & 19.88 \\
Qwen3.5-2B-Base & 2B & 37.98 & 38.59 & 18.79 & 35.12 & 19.74 & 36.38 & 2.92 & 84.80 \\
Qwen3.5-2B & 2B & 35.41 & 35.86 & 19.86 & 36.79 & 20.57 & 37.56 & 2.78 & 32.75 \\
Qwen3.5-4B-Base & 4B & 63.76 & 64.75 & 26.87 & 43.07 & 30.08 & 45.51 & 29.14 & 86.55 \\
Qwen3.5-4B & 4B & 66.49 & 67.25 & 24.38 & 41.53 & 25.56 & 42.79 & 28.52 & 74.27 \\
Qwen3.5-9B-Base & 9B & 69.14 & 69.83 & 30.92 & 48.43 & 35.79 & 52.06 & 47.66 & 86.55 \\
Qwen3.5-9B & 9B & 71.57 & 72.55 & 28.18 & 46.20 & 29.73 & 47.46 & 46.54 & 84.21 \\
Qwen3.5-27B & 27B & 30.33 & 31.92 & 21.76 & 39.23 & 22.24 & 39.65 & 40.97 & 59.06 \\
Qwen3.5-35B-A3B-Base & 35B-A3B & 76.04 & 77.71 & 32.94 & 50.38 & 38.41 & 54.15 & 49.70 & 84.80 \\
Qwen3.5-35B-A3B & 35B-A3B & 79.61 & 80.67 & 22.59 & 40.21 & 23.07 & 40.63 & {\color[HTML]{FE0000} 62.51} & 91.81 \\ \midrule
Gemma-4-E2B & 5B & 17.97 & 18.12 & 15.34 & 29.97 & 15.93 & 30.73 & -17.84 & {\color[HTML]{FE0000} 100.00} \\
Gemma-4-E2B-it & 5B & 5.91 & 10.01 & 4.76 & 10.94 & 4.88 & 11.01 & 13.17 & 85.38 \\
Gemma-4-E4B & 8B & 47.38 & 48.07 & 18.43 & 32.40 & 19.38 & 33.03 & 24.11 & 84.80 \\
Gemma-4-E4B-it & 8B & 58.61 & 54.21 & 12.01 & 21.05 & 12.13 & 21.18 & 9.95 & 39.77 \\
Gemma-4-26B-A4B & 26B-A4B & 58.53 & 59.36 & 17.12 & 31.43 & 18.43 & 33.31 & 16.49 & 92.98 \\
Gemma-4-26B-A4B-it & 26B-A4B & 29.95 & 31.08 & 15.58 & 29.13 & 16.65 & 29.97 & 19.02 & 42.69 \\
Gemma-4-31B & 31B & 74.30 & 75.36 & 19.86 & 35.12 & 21.17 & 36.86 & 42.90 & 88.30 \\
Gemma-4-31B-it & 31B & 42.30 & 44.28 & 13.79 & 26.41 & 13.91 & 26.48 & 9.95 & 30.41 \\ \midrule \midrule
\multicolumn{2}{c}{\textbf{Average}} & 49.91 & 50.67 & 21.61 & 35.76 & 23.66 & 37.63 & 26.14 & 73.48 \\ \midrule
\multicolumn{2}{c}{\textbf{Average (chat-template subset)}} & 46.45 & 47.26 & 22.12 & 35.84 & 24.13 & 37.64 & 28.42 & 65.14 \\ \bottomrule
\end{tabular}%
}
\caption{Open-weight generation results under plain prompting. The chat-template subset average uses the same model subset as the chat-template prompting results in Table~\ref{tab:generation_chat}. The highest score in each metric column is highlighted in red.}
\label{tab:generation_plain}
\end{table*}
\begin{table*}[htbp]
\centering
\resizebox{\textwidth}{!}{%
\begin{tabular}{@{}lrrrrrrrrr@{}}
\toprule
\multicolumn{1}{c}{} & \multicolumn{1}{c}{} & \multicolumn{2}{c}{\textbf{Ko-GSM8K}} & \multicolumn{4}{c}{\textbf{Ko-IFEval}} & \multicolumn{2}{c}{\textbf{Ko-EQ-Bench}} \\ \cmidrule(l){3-4}\cmidrule(l){5-8}\cmidrule(l){9-10}
\multicolumn{1}{c}{\multirow{-2}{*}{\textbf{Model}}} & \multicolumn{1}{c}{\multirow{-2}{*}{\textbf{\begin{tabular}[c]{@{}c@{}}Parameter \\ size\end{tabular}}}} & \multicolumn{1}{c}{\textbf{\begin{tabular}[c]{@{}c@{}}strict\\ -match\end{tabular}}} & \multicolumn{1}{c}{\textbf{\begin{tabular}[c]{@{}c@{}}flexible\\ -extract\end{tabular}}} & \multicolumn{1}{c}{\textbf{\begin{tabular}[c]{@{}c@{}}prompt\\ \_level\\ \_strict\\ \_acc\end{tabular}}} & \multicolumn{1}{c}{\textbf{\begin{tabular}[c]{@{}c@{}}inst\\ \_level\\ \_strict\\ \_acc\end{tabular}}} & \multicolumn{1}{c}{\textbf{\begin{tabular}[c]{@{}c@{}}prompt\\ \_level\\ \_loose\\ \_acc\end{tabular}}} & \multicolumn{1}{c}{\textbf{\begin{tabular}[c]{@{}c@{}}inst\\ \_level\\ \_loose\\ \_acc\end{tabular}}} & \multicolumn{1}{c}{\textbf{eqbench}} & \multicolumn{1}{c}{\textbf{\begin{tabular}[c]{@{}c@{}}percent\\ \_parseable\end{tabular}}} \\ \midrule \midrule
HyperCLOVAX-SEED-Text-Instruct-0.5B & 0.5B & 3.56 & 6.07 & 17.48 & 32.33 & 18.79 & 34.49 & 25.99 & 95.91 \\
HyperCLOVAX-SEED-Text-Instruct-1.5B & 1.5B & 12.51 & 11.90 & 22.24 & 36.31 & 24.38 & 39.23 & 41.32 & 93.57 \\ \midrule
A.X 4.0 Light & 7B & 82.18 & 82.34 & 73.72 & 83.07 & 76.10 & 85.09 & 56.04 & {\color[HTML]{FE0000} 100.00} \\ \midrule
Mi:dm 2.0 Mini & 2.3B & 66.03 & 67.32 & 53.98 & 65.78 & 57.67 & 68.43 & 16.30 & 46.20 \\
Mi:dm 2.0 Base & 11.5B & 32.60 & 84.08 & 70.15 & 80.63 & 74.79 & 83.76 & 47.32 & 78.95 \\ \midrule
EXAONE-4.0-1.2B & 1.2B & 46.10 & 47.92 & 50.30 & 61.53 & 53.27 & 63.76 & 22.36 & 77.78 \\
EXAONE-4.0-32B & 32B & 70.51 & 76.27 & 74.79 & 83.69 & 80.62 & 87.60 & 61.06 & {\color[HTML]{FE0000} 100.00} \\ \midrule
Kanana-2-30b-a3b-instruct-2601 & 30B-A3B & 44.88 & 86.35 & 76.58 & 84.95 & 81.45 & 88.15 & 59.31 & 98.25 \\ \midrule
Qwen3.5-0.8B & 0.8B & 23.05 & 24.41 & 32.46 & 48.43 & 34.13 & 50.38 & 16.77 & 99.42 \\
Qwen3.5-2B & 2B & 43.06 & 44.35 & 44.59 & 61.18 & 48.75 & 64.46 & 31.16 & {\color[HTML]{FE0000} 100.00} \\
Qwen3.5-4B & 4B & 0.15 & 2.12 & 20.93 & 38.33 & 21.52 & 38.82 & 0.00 & 0.00 \\
Qwen3.5-9B & 9B & 0.00 & 2.73 & 20.57 & 38.12 & 20.81 & 38.54 & 0.00 & 0.00 \\
Qwen3.5-27B & 27B & 0.00 & 2.73 & 21.05 & 38.68 & 21.40 & 38.95 & 0.00 & 0.00 \\
Qwen3.5-35B-A3B & 35B-A3B & 0.08 & 3.26 & 21.17 & 38.82 & 21.52 & 39.16 & 0.00 & 0.00 \\
Gemma-4-E2B-it & 5B & 33.59 & 54.13 & 66.83 & 78.54 & 72.06 & 82.44 & 53.64 & {\color[HTML]{FE0000} 100.00} \\
Gemma-4-E4B-it & 8B & 63.76 & 67.55 & 74.79 & 84.18 & 78.72 & 86.97 & 68.86 & {\color[HTML]{FE0000} 100.00} \\
Gemma-4-26B-A4B-it & 26B-A4B & 77.86 & 80.67 & 80.50 & 87.74 & 83.35 & 89.90 & 69.14 & {\color[HTML]{FE0000} 100.00} \\
Gemma-4-31B-it & 31B & {\color[HTML]{FE0000} 82.71} & {\color[HTML]{FE0000} 84.38} & {\color[HTML]{FE0000} 85.02} & {\color[HTML]{FE0000} 90.24} & {\color[HTML]{FE0000} 87.75} & {\color[HTML]{FE0000} 92.26} & {\color[HTML]{FE0000} 73.10} & {\color[HTML]{FE0000} 100.00} \\ \midrule \midrule
\multicolumn{2}{c}{\textbf{Average}} & 37.92 & 46.03 & 50.40 & 62.92 & 53.17 & 65.13 & 35.69 & 71.67 \\ \bottomrule
\end{tabular}%
}
\caption{Open-weight generation results under chat-template prompting. The highest score in each metric column is highlighted in red.}
\label{tab:generation_chat}
\end{table*}
\begin{table*}[htbp]
\centering
\resizebox{\textwidth}{!}{%
\begin{tabular}{@{}lrrrrrrrrrrrr@{}}
\toprule
\multicolumn{1}{c}{\multirow{2}{*}{\textbf{Model}}} & \multicolumn{1}{c}{\textbf{\begin{tabular}[c]{@{}c@{}}Ko\\ -WinoGrande\\ -gen\end{tabular}}} & \multicolumn{1}{c}{\textbf{\begin{tabular}[c]{@{}c@{}}Ko\\ -LAMBADA\\ -gen\end{tabular}}} & \multicolumn{1}{c}{\textbf{\begin{tabular}[c]{@{}c@{}}Ko-ARC\\ -Easy\\ -gen\end{tabular}}} & \multicolumn{1}{c}{\textbf{\begin{tabular}[c]{@{}c@{}}Ko-ARC\\ -Challenge\\ -gen\end{tabular}}} & \multicolumn{2}{c}{\textbf{Ko-GSM8K}} & \multicolumn{4}{c}{\textbf{Ko-IFEval}} & \multicolumn{2}{c}{\textbf{Ko-EQ-Bench}} \\ \cmidrule(l){2-5}\cmidrule(l){6-7}\cmidrule(l){8-11}\cmidrule(l){12-13}
\multicolumn{1}{c}{} & \multicolumn{1}{c}{\textbf{\begin{tabular}[c]{@{}c@{}}exact\\ \_match\end{tabular}}} & \multicolumn{1}{c}{\textbf{\begin{tabular}[c]{@{}c@{}}exact\\ \_match\end{tabular}}} & \multicolumn{1}{c}{\textbf{\begin{tabular}[c]{@{}c@{}}exact\\ \_match\end{tabular}}} & \multicolumn{1}{c}{\textbf{\begin{tabular}[c]{@{}c@{}}exact\\ \_match\end{tabular}}} & \multicolumn{1}{c}{\textbf{\begin{tabular}[c]{@{}c@{}}strict\\ \_match\end{tabular}}} & \multicolumn{1}{c}{\textbf{\begin{tabular}[c]{@{}c@{}}flexible\\ \_extract\end{tabular}}} & \multicolumn{1}{c}{\textbf{\begin{tabular}[c]{@{}c@{}}prompt\\ \_level\\ \_strict\\ \_acc\end{tabular}}} & \multicolumn{1}{c}{\textbf{\begin{tabular}[c]{@{}c@{}}inst\\ \_level\\ \_strict\\ \_acc\end{tabular}}} & \multicolumn{1}{c}{\textbf{\begin{tabular}[c]{@{}c@{}}prompt\\ \_level\\ \_loose\\ \_acc\end{tabular}}} & \multicolumn{1}{c}{\textbf{\begin{tabular}[c]{@{}c@{}}inst\\ \_level\\ \_loose\\ \_acc\end{tabular}}} & \multicolumn{1}{c}{\textbf{eqbench}} & \multicolumn{1}{c}{\textbf{\begin{tabular}[c]{@{}c@{}}percent\\ \_parseable\end{tabular}}} \\ \midrule \midrule
Claude Haiku 4.5 & 77.82 & 91.8 & 97.9 & 94.17 & 93.48 & 93.48 & 71.7 & 82.37 & 81.45 & 88.5 & 65.79 & 98.25 \\
Claude Opus 4.7 & 93.37 & 98.76 & 99.12 & 97.17 & 96.06 & 96.06 & 79.55 & 87.25 & 83.47 & 89.69 & 79.23 & 100 \\ \midrule
GPT-5-mini & 76.95 & 64.66 & 97.69 & 95.89 & 81.96 & 94.47 & 36.98 & 35.75 & 37.1 & 35.82 & 64.9 & 99.42 \\
GPT-5.5 & 91.71 & 96.63 & 98.91 & 97.77 & 96.29 & 96.29 & 81.09 & 83.69 & 82.16 & 84.39 & 78.71 & 100 \\ \bottomrule
\end{tabular}%
}
\caption{Closed API model results across all benchmark tasks under chat-style prompting.}
\label{tab:api_total}
\end{table*}

\begin{table*}[]
\centering
\resizebox{\textwidth}{!}{%
\begin{tabular}{@{}lrrrrr@{}}
\toprule
\multicolumn{1}{c}{} & \multicolumn{1}{c}{} & \multicolumn{1}{c}{\textbf{Ko-WinoGrande-gen}} & \multicolumn{1}{c}{\textbf{Ko-LAMBADA-gen}} & \multicolumn{1}{c}{\textbf{Ko-ARC-Easy-gen}} & \multicolumn{1}{c}{\textbf{Ko-ARC-Challenge-gen}} \\ \cmidrule(l){3-6} 
\multicolumn{1}{c}{\multirow{-2}{*}{\textbf{Model}}} & \multicolumn{1}{c}{\multirow{-2}{*}{\textbf{Parameter size}}} & \multicolumn{1}{c}{\textbf{exact\_match}} & \multicolumn{1}{c}{\textbf{exact\_match}} & \multicolumn{1}{c}{\textbf{exact\_match}} & \multicolumn{1}{c}{\textbf{exact\_match}} \\ \midrule \midrule
HyperCLOVAX-SEED-Text-Instruct-0.5B & 0.5B & 22.18 & 3.90 & 66.71 & 49.44 \\
HyperCLOVAX-SEED-Text-Instruct-1.5B & 1.5B & 48.38 & 61.02 & 76.01 & 58.35 \\ \midrule
A.X 4.0 Light & 7B & 67.25 & 85.90 & 93.22 & 86.38 \\ \midrule
Mi:dm 2.0 Mini & 2.3B & 57.70 & 74.19 & 84.51 & 73.09 \\
Mi:dm 2.0 Base & 11.5B & 68.19 & 86.65 & 94.23 & 87.06 \\ \midrule
EXAONE-4.0-1.2B & 1.2B & 50.59 & 62.84 & 73.61 & 59.30 \\
EXAONE-4.0-32B & 32B & 69.77 & 83.90 & 94.82 & 88.43 \\ \midrule
Kanana-2-30b-a3b-instruct-2601 & 30B-A3B & 64.40 & 76.32 & 95.41 & 88.52 \\ \midrule
Qwen3.5-0.8B & 0.8B & 50.20 & 51.66 & 63.38 & 51.16 \\
Qwen3.5-2B & 2B & 38.44 & 54.86 & 81.40 & 70.27 \\
Qwen3.5-4B & 4B & 0.00 & 0.00 & 0.04 & 0.00 \\
Qwen3.5-9B & 9B & 0.00 & 0.00 & 0.00 & 0.00 \\
Qwen3.5-27B & 27B & 0.00 & 0.00 & 0.00 & 0.00 \\
Qwen3.5-35B-A3B & 35B-A3B & 0.00 & 0.00 & 0.00 & 0.00 \\ \midrule
Gemma-4-E2B-it & 5B & 52.01 & 40.89 & 84.05 & 72.15 \\
Gemma-4-E4B-it & 8B & 65.04 & 81.82 & 93.60 & 86.98 \\
Gemma-4-26B-A4B-it & 26B-A4B & 73.16 & 90.64 & 97.47 & 93.83 \\
Gemma-4-31B-it & 31B & {\color[HTML]{FE0000} 84.85} & {\color[HTML]{FE0000} 96.14} & {\color[HTML]{FE0000} 98.06} & {\color[HTML]{FE0000} 96.57} \\ \midrule \midrule
\multicolumn{2}{c}{\textbf{Average}} & 45.12 & 52.82 & 66.47 & 58.97 \\ \bottomrule
\end{tabular}%
}
\caption{Generative MCQA results for open-weight models under chat-style prompting. The highest score in each metric column is highlighted in red.}
\label{tab:generative_mcqa_chat}
\end{table*}
\begin{table*}[]
\centering
\resizebox{\textwidth}{!}{%
\begin{tabular}{@{}lrrrrrrrrr@{}}
\toprule
\multicolumn{1}{c}{\multirow{2}{*}{\textbf{Model}}} &
  \multicolumn{1}{c}{\multirow{2}{*}{\textbf{Parameter size}}} &
  \multicolumn{1}{c}{\textbf{\begin{tabular}[c]{@{}c@{}}Ko\\ -WinoGrande\\ -gen\end{tabular}}} &
  \multicolumn{1}{c}{\textbf{\begin{tabular}[c]{@{}c@{}}Ko\\ -LAMBADA\\ -gen\end{tabular}}} &
  \multicolumn{1}{c}{\textbf{\begin{tabular}[c]{@{}c@{}}Ko-ARC\\ -Easy\\ -gen\end{tabular}}} &
  \multicolumn{1}{c}{\textbf{\begin{tabular}[c]{@{}c@{}}Ko-ARC\\ -Challenge\\ -gen\end{tabular}}} &
  \multicolumn{2}{c}{\textbf{Ko-GSM8K}} &
  \multicolumn{2}{c}{\textbf{Ko-EQ-Bench}} \\ \cmidrule(l){3-10}  
 &
  \multicolumn{1}{c}{} &
  \multicolumn{1}{c}{\textbf{\begin{tabular}[c]{@{}c@{}}exact\\ \_match\end{tabular}}} &
  \multicolumn{1}{c}{\textbf{\begin{tabular}[c]{@{}c@{}}exact\\ \_match\end{tabular}}} &
  \multicolumn{1}{c}{\textbf{\begin{tabular}[c]{@{}c@{}}exact\\ \_match\end{tabular}}} &
  \multicolumn{1}{c}{\textbf{\begin{tabular}[c]{@{}c@{}}exact\\ \_match\end{tabular}}} &
  \multicolumn{1}{c}{\textbf{\begin{tabular}[c]{@{}c@{}}strict\\ \_match\end{tabular}}} &
  \multicolumn{1}{c}{\textbf{\begin{tabular}[c]{@{}c@{}}flexible\\ \_extract\end{tabular}}} &
  \multicolumn{1}{c}{\textbf{eqbench}} &
  \multicolumn{1}{c}{\textbf{\begin{tabular}[c]{@{}c@{}}percent\\ \_parseable\end{tabular}}} \\ \midrule
Qwen3.5-0.8B    & 0.8B & 50.12 & 51.71 & 62.58 & 50.64 & 23.96 & 25.17 & 16.65 & 99.42  \\
Qwen3.5-2B      & 2B   & 38.20 & 55.17 & 81.57 & 69.75 & 44.50 & 45.34 & 30.98 & 100.00 \\
Qwen3.5-4B      & 4B   & 62.51 & 81.37 & 93.52 & 86.89 & 69.60 & 71.04 & 53.11 & 100.00 \\
Qwen3.5-9B      & 9B   & 68.19 & 87.01 & 96.34 & 91.09 & 77.33 & 78.54 & 54.55 & 100.00 \\
Qwen3.5-27B     & 27B  & 80.58 & 95.92 & 98.53 & 96.06 & 80.67 & 82.87 & 72.47 & 100.00 \\
Qwen3.5-35B-A3B & 35B-A3B  & 74.43 & 91.22 & 97.73 & 94.26 & 80.67 & 82.94 & 70.45 & 100.00 \\ \bottomrule
\end{tabular}%
}
\caption{Generative MCQA and generation results for six instruction-tuned Qwen3.5 models with \texttt{enable\_thinking=False}, while retaining chat templates.}
\label{tab:qwen_nothink}
\end{table*}
\begin{table*}[]
\centering
\resizebox{0.45\textwidth}{!}{%
\begin{tabular}{@{}llrrr@{}}
\toprule
\multicolumn{1}{c}{\textbf{Task}} &
  \multicolumn{1}{c}{\textbf{Model group}} &
  \multicolumn{1}{c}{\textbf{English}} &
  \multicolumn{1}{c}{\textbf{DeepL}} &
  \multicolumn{1}{c}{\textbf{KBS}} \\ \midrule
\multirow{3}{*}{ARC}        & Korean-developed & 67.29 & 54.00 & 59.49 \\
                            & Qwen3.5          & 51.22 & 44.14 & 46.49 \\
                            & Gemma 4          & 72.56 & 60.14 & 67.05 \\ \midrule
\multirow{3}{*}{GSM8K}      & Korean-developed & 57.52 & 34.36 & 44.80 \\
                            & Qwen3.5          & 54.11 & 10.01 & 11.06 \\
                            & Gemma 4          & 82.41 & 59.84 & 64.48 \\ \midrule
\multirow{3}{*}{WinoGrande} & Korean-developed & 57.52 &       & 56.49 \\
                            & Qwen3.5          & 55.32 &       & 54.54 \\
                            & Gemma 4          & 55.98 &       & 55.98 \\ \midrule
\multirow{3}{*}{EQ-Bench}   & Korean-developed & 54.48 &       & 41.21 \\
                            & Qwen3.5          & 13.63 &       & 7.99  \\
                            & Gemma 4          & 75.03 &       & 66.19 \\ \bottomrule
\end{tabular}%
}
\caption{Performance across English, raw DeepL, and KBS variants under chat-style prompting, averaged over the same eight Korean-developed, six Qwen3.5, and four Gemma 4 models. ARC and WinoGrande use NPSQ accuracy, GSM8K uses strict exact match, and EQ-Bench uses its benchmark score. DeepL results are unavailable for WinoGrande and excluded for EQ-Bench because the Korean emotion labels in the translated prompts do not match the English reference labels used by the scorer.}
\label{tab:language_comparison_appendix}
\end{table*}

\end{document}